\documentclass{article}
\usepackage{arxiv}

\usepackage[utf8]{inputenc} 
\usepackage[T1]{fontenc}    
\usepackage{hyperref}       
\usepackage{url}            
\usepackage{booktabs}       
\usepackage{amsfonts}       
\usepackage{nicefrac}       
\usepackage{microtype}      
\usepackage{lipsum}		
\usepackage{subcaption}
\usepackage{graphicx}
\usepackage{natbib}
\usepackage{doi}
\usepackage{amsmath}
\usepackage{float}
\usepackage{wrapfig}

\title{Fourier-Invertible Neural Encoder (FINE) for Homogeneous Flows}

\date{May 21, 2025}	

\author{
    \href{https://orcid.org/0009-0004-8771-3616}
    {\includegraphics[scale=0.06]{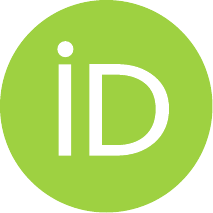}\hspace{1mm}Anqiao Ouyang} \\
        Mount Boucherie Secondary School \\
        West Kelowna, British Columbia, Canada \\\\
	Department of Aerospace Engineering \\
	San Diego State University\\
	San Diego, California, USA \\
	\And
	\href{https://orcid.org/0009-0009-3997-8157}{\includegraphics[scale=0.06]{orcid.pdf}\hspace{1mm}Hongyi Ke} \\
	Department of Aerospace Engineering\\
	San Diego State University\\
	San Diego, California, USA \\
        \And
        \href{https://orcid.org/0000-0001-7393-9922}
    {\includegraphics[scale=0.06]{orcid.pdf}\hspace{1mm}Qi Wang} \\
	Department of Aerospace Engineering\\
	San Diego State University\\
	San Diego, California, USA \\
	\texttt{qwang4@sdsu.edu} \\
}
\renewcommand{\shorttitle}{Fourier-Invertible Neural Encoder (FINE) for Homogeneous Flows}

\hypersetup{
pdftitle={Fourier-Invertible Neural Encoder (FINE) for Homogeneous Flows},
pdfsubject={},
pdfauthor={Anqiao Ouyang, Hongyi Ke, Qi Wang},
pdfkeywords={Scientific Machine learning, Auto Encoder, Dimension Reduction},
}

\begin{document}
\maketitle

\begin{abstract}
We present the Fourier-Invertible Neural Encoder (FINE), a compact and interpretable architecture for dimension reduction in translation-equivariant datasets. FINE integrates reversible filters and monotonic activation functions with a Fourier truncation bottleneck, achieving information-preserving compression that respects translational symmetry. This design offers a new perspective on symmetry-aware learning, linking spectral truncation to group-equivariant representations.
The proposed FINE architecture is tested on one-dimensional nonlinear wave interaction, one-dimensional Kuramoto–Sivashinsky turbulence dataset, and a two-dimensional turbulence dataset. FINE achieves an overall 4.9–9.1x lower reconstruction error than convolutional autoencoders while using only 13–21\% of their parameters. The results highlight FINE's effectiveness in representing complex physical systems with minimal dimension in the latent space.
The proposed framework provides a principled framework for interpretable, low-parameter, and symmetry-preserving dimensional reduction, bridging the gap between Fourier representations and modern neural architectures for scientific and physics-informed learning.
\end{abstract} 

\section{Introduction}
\label{sec:intro}
Understanding complex patterns in physics-based simulation data, such as turbulence, oceanographic flows, and astrophysics, remains a fundamental challenge in scientific machine learning (Sci-ML). The intricate nonlinear interactions and symmetries must be captured through dimension reduction. In canonical flow systems, such as three-dimensional isotropic turbulence \citep{park2025coherent} and wall-bounded turbulence \citep{motoori2021hierarchy}, coherent structures and spatial correlations suggest the existence of low-dimensional manifolds that encode essential physical behavior. 
These datasets often contain symmetries crucial for reduced-order models, such as circular or translational equivariance. Harmonic Networks (H-Nets) achieve equivariance to translations and rotations by employing circular harmonics as filters, enabling consistent responses to rotated inputs without requiring multiple rotated versions of the filters \citep{worrall2017harmonic}. Additionally, Tensor Field Networks are locally equivariant to 3D rotations and translations, utilizing filters built from spherical harmonics to process 3D point clouds effectively \citep{Thomas2018}.
Renormalization group (RG) symmetry also plays a significant role in understanding turbulence. Onsager's ``ideal turbulence" theory describes turbulent energy cascades without probabilistic assumptions and yields a local, deterministic version of the Kolmogorov 4/5th law, offering insights into energy transfer mechanisms in turbulent flows \citep{eyink2024onsager}. Furthermore, the transfer of kinetic energy from large to small scales in turbulence is commonly attributed to the stretching of vorticity by the strain rate, but strain self-amplification also plays a role, enhancing the understanding of turbulence dynamics \citep{Johnson2020}.
Deep learning-based autoencoders have achieved great success in learning data-driven low-dimensional manifolds. For example, extreme aerodynamic flows can be compressed through machine learning into a low-dimensional manifold, enabling real-time sparse reconstruction and control of extremely unsteady gusty flows \citep{FukamiTaira2023}. Expanding upon the discussion of deep learning-based autoencoders in fluid dynamics, recent advancements have demonstrated that effective learning of turbulent flow structures can be achieved even with minimal data. This method leverages the inherent scale-invariance and multiscale features present in turbulence, enabling the model to reconstruct high-resolution flow fields from low-resolution inputs across various Reynolds numbers. \citep{fukami2024single}.
Despite these achievements, most existing autoencoders do not respect these innate symmetries of physics data.
In addition, the network designs are inherited from computer vision, relying on fixed activation functions and non-invertible structures. These architectures also prioritize expressivity over interpretability or physical consistency. 
The current work introduces a novel framework for physics-based representation learning: Fourier-Invertible Neural Encoder (FINE). Built upon flexible, monotonous activation functions and invertible filters. This approach integrates invertible single-neuron layers with a Fourier truncation bottleneck to enforce circular shift equivariance and enable dimension reduction. 

\subsection{Homogeneous Flow Fields and Dimension Reduction}
Many physical systems exhibit spatial homogeneity, where the underlying dynamics remain invariant under spatial translations. In such cases, the dataset $\mathcal{D}$ satisfies $\mathcal{T}\mathcal{D}=\mathcal{D}$, where $\mathcal{T}$ denotes a translation operator~\citep{helwig2023group}. Translation-equivariant structures are particularly prominent in fluid dynamics, where homogeneous turbulence, channel flows, and periodic shear flows arise in idealized or canonical settings. The importance of such homogeneous flows and their coherent structures is extensively reviewed in wall-bounded turbulence~\citep{jimenez2018coherent}, which highlights the hierarchical organization of self-similar motions that underpin energy transfer and scale interactions.

The large-scale nature of turbulent flow datasets imposes severe computational costs, motivating the use of dimension reduction. Classical linear methods such as Proper Orthogonal Decomposition (POD)~\citep{lumley1967structure}, Principal Component Analysis (PCA)~\citep{abdi2010principal}, Dynamic Mode Decomposition (DMD)~\citep{schmid2010dynamic}, and Spectral POD (SPOD)~\citep{towne2018spectral} remain widely used. These methods, derived from the Singular Value Decomposition (SVD), extract spatial and temporal modes that optimally capture energy or spectral content. For translation-invariant datasets, the eigenfunctions of the covariance operator correspond to Fourier modes~\citep{bolla2021block}, reflecting the inherent cyclic symmetry of homogeneous systems. However, these linear decompositions are limited by their assumption of mode independence and linear superposition, an assumption violated by the strongly coupled nonlinear interactions in turbulence. 
The limitations of linear reduction techniques have motivated the development of nonlinear approaches, particularly \emph{autoencoders} (AEs), which learn low-dimensional latent manifolds through data-driven nonlinear mappings~\citep{hinton2006reducing}.  \citet{khoa2023convolutional} demonstrated that convolutional autoencoders can reconstruct near-wall turbulence and rare events more accurately than POD or DMD. \citet{fukami2019super} used deep super-resolution autoencoders to upsample coarse flow data to high fidelity, and \citet{FukamiTaira2023} showed that CAE-based manifolds can be used for real-time control of unsteady gusts. 

The proposed FINE architecture builds on these ideas of nonlinear decomposition and achieves dimension reduction as a form of \emph{equivariant information compression}, a reversible transformation followed by a controlled Fourier truncation that preserves translation symmetry. This view aligns with the mathematical foundations of homogeneous turbulence, where energy and structure are distributed across invariant wavenumbers. By integrating invertible layers with spectral projection, FINE bridges traditional Fourier-based representations and modern symmetry-aware neural encoders.

\subsection{Group-Equivariant Deep Learning}
Standard CAEs do not explicitly encode physical symmetries such as translation or rotation invariance; the symmetries in the data are often dealt with by data augmentation. Their learned representations can therefore depend on the spatial placement of flow features, limiting interpretability and generalization across shifted or rotated domains.

To overcome this deficiency, the past decade has seen rapid progress in the development of \textit{group-equivariant neural networks}, which integrate symmetry constraints directly into the network structure. Group-Equivariant Convolutional Neural Networks (G-CNNs)~\citep{cohen2016group} first formalized the idea of convolutions over general symmetry groups. By convolving over both spatial coordinates and group elements, G-CNNs guarantee that if an input is transformed by a symmetry operation (e.g., rotation or reflection), the resulting feature maps transform predictably under the same group. This architecture achieves exact equivariance to discrete symmetry groups (such as planar rotations and reflections) and demonstrates improved data efficiency and generalization for tasks where geometric invariance is essential.
Harmonic Networks (H-Nets)~\citep{worrall2017harmonic} extended this principle to continuous rotations using complex-valued circular harmonics as convolutional filters. This approach achieves continuous rotation equivariance in two dimensions without the need for multiple filter orientations, making it suitable for physical fields defined over periodic domains such as vorticity or scalar concentration. Building on this foundation, Tensor Field Networks (TFNs)~\citep{Thomas2018} generalized equivariance to three-dimensional Euclidean transformations by encoding features as geometric tensors that transform according to irreducible representations of SO(3). TFNs, and later SE(3)-Transformers~\citep{fuchs2020se}, became the standard tools for modeling 3D physical and molecular systems, achieving state-of-the-art accuracy in molecular force prediction and point-cloud processing while preserving physical consistency.
Subsequent developments such as Gauge-Equivariant CNNs~\citep{cohen2019gauge} and E(3)-Equivariant Graph Neural Networks (EGNNs)~\citep{satorras2021n} expanded these ideas to curved manifolds and irregular graphs, enabling equivariant modeling on general non-Euclidean domains. Gauge-equivariant frameworks introduced a formalism to express local symmetry transformations that vary across the input domain, ensuring consistent feature transformation under local coordinate changes: a key concept for modeling complex surfaces and flow geometries. EGNNs, by operating on particle or node positions directly, have found extensive use in atomistic and fluid simulations, achieving impressive results in learning invariant energy functions and particle dynamics. Together, these architectures demonstrate that embedding group symmetries within neural network layers not only improves generalization and stability but also reduces parameter counts and training data requirements—an important property for data-scarce physical modeling.

Parallel to these developments, a different stream of research, namely \textit{operator-learning networks}, has tackled translation-equivariant learning from the perspective of functional mappings. Fourier Neural Operators (FNOs)~\citep{li2020fourier} leverage spectral convolution to learn mappings between function spaces while preserving translation invariance through the Fast Fourier Transform (FFT). By learning the nonlinear coupling among Fourier modes, FNOs efficiently solve parametric PDEs such as the Navier–Stokes equations and Burgers’ equation. Geo-FNO~\citep{Li2024} further extends this framework to irregular geometries by learning coordinate deformations that render the input field locally translation-compatible. These architectures underscore the benefits of spectral-domain learning for homogeneous or periodic physical systems. However, both FNOs and other operator-learning models are designed for \emph{forward} mapping between functions, not for \emph{representation learning} or data-driven compression. They preserve symmetry during prediction, but do not yield an interpretable latent space.

In the context of homogeneous flow fields, preserving translational invariance in the encoding process is crucial, as it ensures that the latent representation reflects only the intrinsic dynamics of the system, independent of spatial positioning. Several recent studies have attempted to address this. For example, Wang et al.~\citep{wang2020towards} incorporated rotationally equivariant convolutions into autoencoders for flow reconstruction, improving generalization across orientations. Other works have explored spectral autoencoders that encode physical fields in the Fourier domain to ensure shift equivariance~\citep{li2023fourier}. Nevertheless, most symmetry-aware models remain \emph{non-invertible}, making it difficult to recover information flow and analyze latent variables in physical terms.
These developments naturally motivate more advanced architectures such as Variational Autoencoders (VAEs) and Koopman Autoencoders (KAEs), which embed additional structure in the latent space. VAEs~\citep{KingmaWelling2014} extend standard autoencoders with probabilistic inference, modeling latent variables as distributions to capture the stochasticity of physical systems and enable generative sampling. Koopman autoencoders~\citep{Azencot2020} combine neural encoders with operator-theoretic principles, mapping nonlinear dynamics into a latent space where they evolve linearly under a learned Koopman operator.

\subsection{Invertible Neural Networks}

Conventional autoencoder architectures learn separate encoder and decoder mappings that are optimized jointly but operate independently. While this separation facilitates training flexibility, it typically sacrifices interpretability and strict information preservation. Once the encoder projects input data to a low-dimensional latent space, some information is irreversibly lost, and there is no guarantee that the decoder’s inverse mapping exactly recovers the input. For physical systems governed by conservation laws or reversible dynamics, such non-invertible architectures pose limitations: they obscure cause–and–effect relationships, disrupt energy consistency, and hinder the interpretability of latent representations.

Invertible neural networks (INNs) address these issues by enforcing bijective transformations throughout the model. Each layer in an INN defines a one-to-one mapping $\boldsymbol{z} = \phi(\boldsymbol{f})$ such that every input can be uniquely inverted via $\boldsymbol{f} = \phi^{-1}(\boldsymbol{z})$. This property guarantees exact information flow, making the representation fully reversible and well-suited for applications in scientific and physical modeling. Early progress in this direction began with Nonlinear Independent Components Estimation (NICE)~\citep{DinhKB14}, which introduced additive coupling layers that ensured tractable inverses and Jacobian determinants. The RealNVP model~\citep{DinhSB17} extended this framework to affine transformations, enabling exact log-likelihood estimation for generative modeling. Subsequently, Glow~\citep{kingma2018glow} refined the architecture using invertible $1\times1$ convolutions, demonstrating scalable bijective image transformations. These models established a new paradigm where deep networks could be exactly inverted without compromising expressive power.

For continuous dynamical systems, invertibility has been extended via the Invertible Residual Network (I-ResNet)~\citep{behrmann2019invertible}. By constraining the Lipschitz constant of each residual block to be less than one, I-ResNets guarantee contraction mappings that are invertible through fixed-point iteration. Spectral normalization or Jacobian regularization is typically applied to enforce stability and tractable Jacobian computation. These methods allow the construction of deep, high-dimensional invertible architectures that maintain stable gradients and preserve all input information. Such invertible models have recently been adopted for physically interpretable learning, such as the energy-conserving reversible networks for Hamiltonian dynamics~\citep{toth2019hamiltonian}. Across these domains, invertibility has been shown to improve numerical stability, facilitate uncertainty quantification, and yield physically consistent reconstructions.

Despite these advances, most existing invertible networks have been developed in the context of density estimation or generative modeling rather than dimension reduction. In addition, they generally operate in spatial or pixel space, without incorporating known spectral symmetries of physical systems. This limits their interpretability when applied to homogeneous or translation-equivariant datasets, where dynamics are more naturally expressed in the Fourier domain. 

The proposed Fourier-Invertible Neural Encoder (FINE) addresses this gap by introducing a structured, invertible mapping that operates in both spatial and spectral spaces. FINE combines strictly monotonic activation functions with invertible filter layers, ensuring bijectivity throughout the encoding and decoding process, while applying a Fourier truncation at the latent bottleneck to perform controlled dimensional reduction. This design ensures that all information loss is explicit and interpretable, arising solely from the truncation of high-frequency spectral components.
The remainder of this paper is organized as follows. Section~\ref{sec:method} presents the mathematical framework and implementation of FINE. Sections~\ref{sec:Toy1} and~\ref{sec:ks} demonstrate its performance on nonlinear wave interactions and one-dimensional turbulence governed by the Kuramoto–Sivashinsky equation. Comparative analyses with CNN-based autoencoders highlight improvements in accuracy, parameter efficiency, and physical interpretability.

\section{Fourier-Invertible Neural Encoder (FINE) Architecture}
\label{sec:method}

We consider the problem of learning a \emph{translation-equivariant}, \emph{invertible} operator acting on functions $f : \Omega \to \mathbb{R}$, where the spatial domain is multi-periodic, i.e., $\Omega = [0, L_1) \times [0, L_2) \times \cdots \times [0, L_d) = [0, \boldsymbol{L})^d$. 
The \emph{circular translation operator} \( \mathcal{T}_{\boldsymbol{a}} \) acts on \( f \) as
\[
(\mathcal{T}_{\boldsymbol{a}} f)(\boldsymbol{x})
= f\big((\boldsymbol{x} - \boldsymbol{a}) \bmod \boldsymbol{L}\big),
\]
where the modulo operation is applied componentwise with respect to the domain lengths
\( \boldsymbol{L} = (L_1, \dots, L_d) \). Without inviting ambiguity, we just write $f(\boldsymbol{x} - \boldsymbol{a})$ as shorthand.

An operator \( \mathcal{G} \) is said to be \emph{translation-equivariant} if it commutes with every translation, i.e.,
\[
\mathcal{G}\big[\mathcal{T}_{\boldsymbol{a}} f\big]
= \mathcal{T}_{\boldsymbol{a}}\big[\mathcal{G} f\big],
\quad \forall\, \boldsymbol{a} \in \Omega.
\]

This property is fundamental in homogeneous or periodic systems, ensuring that the learned representation is independent of the absolute spatial location of features.

In practice, we represent \( f \) by its finite-dimensional discretization
\[
\mathbf{f} = \bigl(f(\boldsymbol{x}_1), f(\boldsymbol{x}_2), \ldots, f(\boldsymbol{x}_N)\bigr)^\top \in \mathbb{R}^N,
\]
where \( \boldsymbol{x}_j \in \Omega \subset \mathbb{R}^d \) denote uniformly spaced grid points
in the multi-periodic domain.
A spatial translation of the continuous function \( f \) then corresponds to a
\emph{cyclic permutation} of the entries of \( \mathbf{f} \) along each periodic direction.
Hence, translation equivariance in function space reduces to
\emph{cyclic permutation equivariance} in the discrete representation.

We denote the encoder and decoder as two bijective operators,
\[
\mathcal{G}_e: \mathcal{X}(\Omega) \to \mathcal{Z}, 
\quad \text{and} \quad 
\mathcal{G}_d: \mathcal{Z} \to \mathcal{X}(\Omega),
\]
where \( \mathcal{X}(\Omega) \) is the function space of interest and \( \mathcal{Z} \) is the latent space. 
Their discrete counterparts are denoted by \( \boldsymbol{g}_e \) and \( \boldsymbol{g}_d \), respectively.

\begin{figure*}[t]
    \centering
    \includegraphics[width=\textwidth]{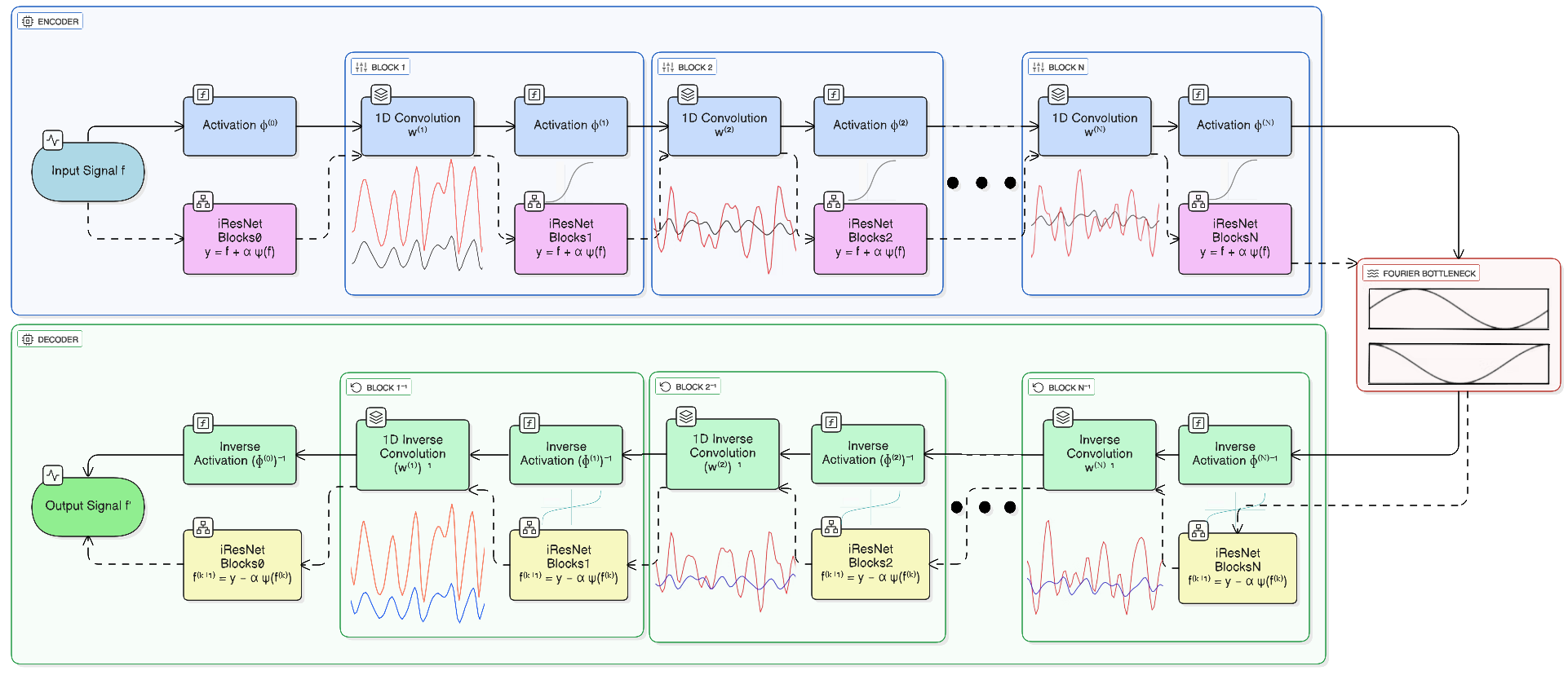}
    \caption{Schematic of the FINE architecture. The encoder consists of a cascade of invertible convolutions and monotonic nonlinearities, followed by a Fourier truncation bottleneck that performs interpretable, symmetry-preserving dimension reduction.}
    \label{fig:schematic}
\end{figure*}

We adopt the universal approximation theorem~\citep{yu2021arbitrary} for translation-equivariant operators; we assume such operators can be approximated by a composition of:
(a) linear convolutions \( \mathcal{W} \), and 
(b) pointwise nonlinearities \( \phi \) applied identically across space. 
These conditions ensure that the network remains equivariant under translation.

The encoder structure of FINE can therefore be expressed as a composition of invertible convolutional and activation operators, followed by a Fourier truncation:
\begin{equation}
    \mathcal{G}_{e}[f]
    = \mathcal{P}_{\mathcal{F}}\,
      \phi^{(L)} \!\circ\! 
      \mathcal{W}^{(L)} \!\circ\! 
      \phi^{(L-1)} \!\circ\! 
      \mathcal{W}^{(L-1)} \!\circ \cdots \circ\!
      \phi^{(1)} \!\circ\!
      \mathcal{W}^{(1)}[f],
    \label{eq:encoder}
\end{equation}
where \( \mathcal{P}_{\mathcal{F}} \) denotes the \emph{Fourier projection operator} that truncates high-wavenumber modes to perform controlled dimensional reduction while preserving translation equivariance.

The corresponding decoder reconstructs the function by inverting each layer sequentially:
\begin{equation}
\begin{aligned}
    \mathcal{G}_{d}[z]
    = &(\mathcal{W}^{(1)})^{-1} \!\circ\! (\phi^{(1)})^{-1} 
      \!\circ\! (\mathcal{W}^{(2)})^{-1} \!\circ\! (\phi^{(2)})^{-1} \!\circ\! 
      \\
      &\cdots 
      \!\circ\! (\mathcal{W}^{(L)})^{-1} \!\circ\! (\phi^{(L)})^{-1}[z].
    \label{eq:decoder}
\end{aligned}
\end{equation}

Equations~(\ref{eq:encoder})–(\ref{eq:decoder}) define a fully invertible mapping between the input function and its latent representation, apart from the explicit information loss introduced by Fourier truncation. 
Invertibility requires two key conditions: (1) the nonlinearities \( \phi^{(l)} \) must be strictly monotonic to admit well-defined inverses, and (2) the convolutional filters \( \mathcal{W}^{(l)} \) must be spectrally invertible, meaning that their Fourier coefficients \( \widehat{w}^{(l)}({\boldsymbol{k}}) \) are nonzero for all retained wavenumbers \( {\boldsymbol{k}} \). 
The overall structure of FINE is shown in Figure \ref{fig:schematic}.
Under these constraints, FINE provides an information-preserving, translation-equivariant, and interpretable mechanism for dimension reduction in function space.
\subsection{Monotonic and Invertible Activations}
\label{sec:activation}

Invertibility of nonlinear activations is essential for maintaining a bijective mapping between the encoder and decoder in FINE.  
Let \( \phi : \mathbb{R} \to \mathbb{R} \) denote a scalar activation function, and let the induced operator \( \Phi : \mathcal{X}(\Omega) \to \mathcal{X}(\Omega) \) act pointwise on functions as
\[
(\Phi f)({\boldsymbol{x}}) = \phi(f({\boldsymbol{x}})).
\]
To ensure that \( \Phi \) is invertible, \( \phi \) must be \emph{strictly monotonic} so that its inverse \( \phi^{-1} \) exists and is well defined for all values in the range of \( \phi \).  
Because the operator acts identically at each spatial location, translation equivariance is automatically preserved:
\[
\Phi[\mathcal{T}_{\boldsymbol{a}} f]({\boldsymbol{x}})
 = \phi\big(f({\boldsymbol{x}} - {\boldsymbol{a}})\big)
 = (\mathcal{T}_{\boldsymbol{a}}[\Phi f])({\boldsymbol{x}}),
 \quad \forall {\boldsymbol{a}} \in \Omega.
\]

We now describe two monotonic activation families that satisfy these requirements and are used in the FINE framework.  
The first, the SmoothReLU$_\varepsilon$ activation, provides a continuously differentiable approximation to ReLU with explicit analytical invertibility and controlled smoothness.  
The second, the residual-type activation \( \phi_{\mathrm{res}} \), achieves invertibility through a small, Lipschitz-bounded perturbation of the identity mapping, following the Invertible-ResNet formulation.  
Together, these constructions establish a mathematically rigorous foundation for reversible, stable, and symmetry-preserving nonlinear transformations within FINE.  
The choice between them depends on the desired balance between smoothness, computational efficiency, and expressive capacity of the learned representation.

\subsubsection*{(a) Smooth Monotonic Piecewise-Linear Activations}

A simple yet expressive class of invertible activations can be constructed from \emph{monotonic piecewise-linear functions}, which is equivalent to a single layer composed of multiple ReLU activations. We use $p+1$ monotonically increasing sequence between $0$ and $1$ representing control points of the activation function $\left\{(i/p, \displaystyle \sum_{j=1}^i e^{-\beta_i}/\sum_{j=1}^p e^{-\beta_i})\right\}_{i=0}^{p}$. The corresponding layer can be realized by stacking ReLU units with appropriately chosen weights and biases.
To guarantee differentiability and stable gradient propagation, we employ a smooth approximation of the ReLU function, denoted by \(\text{SmoothReLU}_{\varepsilon}\), which transitions continuously from zero to linear behavior over a small interval \([-\varepsilon, \varepsilon]\).  
Define
\begin{equation}
\phi_{\varepsilon}(s) = s \, h\!\left(\frac{s+\varepsilon}{2\varepsilon}\right),
\quad
h(t) = t^{2}(3-2t),
\label{eq:smoothrelu}
\end{equation}
where \(t = (s+\varepsilon)/(2\varepsilon)\) maps \(s \in [-\varepsilon, \varepsilon]\) to \(t \in [0,1]\).  
The cubic Hermite interpolant \(h(t)\) satisfies \(h(0)=0\), \(h(1)=1\), and \(h'(t)>0\) for all \(t\in(0,1)\), ensuring strict monotonicity and continuous differentiability of \(\phi_{\varepsilon}\).  

Outside the transition region, \(\phi_{\varepsilon}(s)=\max(0,s)\), while inside it provides a smooth, strictly increasing interpolation between the two regimes.  
Consequently, \(\phi_{\varepsilon}\) is \(C^{1}\), globally monotonic, and invertible with derivative bounded away from zero.  
The inverse \(\phi_{\varepsilon}^{-1}\) can be computed analytically in the piecewise-linear segments or numerically within the transition region using Newton iteration.  
Because the operator acts pointwise on \(f({\boldsymbol{x}})\), the inverse mapping \((\Phi_{\varepsilon}^{-1}f)(\boldsymbol{x})=\phi_{\varepsilon}^{-1}(f(\boldsymbol{x}))\) exists for all \(f\in \text{Range}(\Phi_{\varepsilon})\).

\subsubsection*{(b) Residual-Type Monotonic Activations (I-ResNet Form)}

An alternative monotonic activation used in FINE is derived from the Invertible Residual Network (I-ResNet) formulation~\citep{behrmann2019invertible}.  
For any scalar input \( s \in \mathbb{R} \), the activation is defined as
\begin{equation}
\phi_{\text{res}}(s) = s + \alpha\,\psi(s),
\label{eq:iresnet}
\end{equation}
where \( \psi: \mathbb{R} \to \mathbb{R} \) is a learnable nonlinear function and \( \alpha > 0 \) is a small scaling coefficient.  
The induced operator acting on the input signal \( f:\Omega \to \mathbb{R} \) is therefore
\begin{equation}
(\Phi_{\text{res}} f)(\boldsymbol{x}) = f(\boldsymbol{x}) + \alpha\, \psi(f(\boldsymbol{x})),
\label{eq:iresnet_operator}
\end{equation}
where \( \psi \) acts pointwise on the signal values.

If \( \psi \) is Lipschitz continuous with constant \( L \), i.e.,
\[
|\psi(s_1) - \psi(s_2)| \le L |s_1 - s_2|, \quad \forall\, s_1,s_2 \in \mathbb{R},
\]
and we choose the scaling parameter satisfying \( \alpha L < 1 \) (Spectral normalization), then \( \phi_{\text{res}} \) is strictly monotonic and hence invertible.  
The inverse activation can be computed by fixed-point iteration:
\begin{equation}
s^{(k+1)} = y - \alpha\,\psi(s^{(k)}), 
\qquad s^{(0)} = y,
\label{eq:fixedpoint}
\end{equation}
which converges under the contraction mapping principle, guaranteed by the Lipschitz constraint \( \alpha L < 1 \). 
\subsection{Invertible Filters}
\label{sec:filters}

The convolutional operators used in FINE must be invertible to ensure that the overall encoder–decoder mapping remains bijective.  
Let \( \mathcal{W}: \mathcal{X}(\Omega) \to \mathcal{X}(\Omega) \) denote a convolution operator acting on an input signal \( f:\Omega \to \mathbb{R} \) as
\[
(\mathcal{W}f)(\boldsymbol{x}) = \int_{\Omega} w(\boldsymbol{x} - \boldsymbol{\xi})\,f(\boldsymbol{\xi})\,d\boldsymbol{\xi},
\]
where \( w:\Omega \to \mathbb{R} \) is the convolution kernel.  
By the convolution theorem, \( \mathcal{W} \) diagonalizes in the Fourier domain:
\[
\widehat{(\mathcal{W}f)}(\boldsymbol{k}) = \widehat{w}(\boldsymbol{k})\,\widehat{f}(\boldsymbol{k}),
\]
where \( \widehat{w}(\boldsymbol{k}) = \mathcal{F}[w](\boldsymbol{k}) \) and \( \widehat{f}(\boldsymbol{k}) = \mathcal{F}[f](\boldsymbol{k}) \) denote the Fourier coefficients of the kernel and the signal, respectively.

The operator \( \mathcal{W} \) is invertible if and only if its frequency response \( \widehat{w}(\boldsymbol{k}) \) is nonzero for all \( k \) in the domain of retained modes, i.e.,
\[
|\widehat{w}(\boldsymbol{k})| > 0, \quad \forall \boldsymbol{k} \in \mathcal{K}_{\text{active}}.
\]
Under this condition, the inverse operator \( \mathcal{W}^{-1} \) exists and acts in the spectral domain as
\[
\widehat{(\mathcal{W}^{-1}f)}(\boldsymbol{k}) = \frac{1}{\widehat{w}(\boldsymbol{k})}\,\widehat{f}(\boldsymbol{k}).
\]
Thus, invertibility reduces to ensuring that the Fourier multiplier \( \widehat{w}(\boldsymbol{k}) \) has no zero entries—equivalently, that the kernel has full spectral support over the retained frequency band.  

In the discrete implementation, we parameterize the filters directly in the frequency domain to guarantee spectral invertibility.  
Let \( \mathbf{f} \in \mathbb{R}^N \) be the discrete sampling of \( f(\boldsymbol{x}) \), and let \( \widehat{\mathbf{f}} = \mathcal{F}(\mathbf{f}) \) denote its discrete Fourier transform.  
The discrete convolution layer is then represented as
\[
\mathbf{z} = \mathcal{F}^{-1}\!\big( e^{\boldsymbol{\theta}} \odot \widehat{\mathbf{f}} \big),
\qquad
\mathbf{f}' = \mathcal{F}^{-1}\!\big( e^{-\boldsymbol{\theta}} \odot \widehat{\mathbf{z}} \big),
\]
where \( e^{\boldsymbol{\theta}} \) denotes the elementwise exponential of the real-valued parameter vector \( \boldsymbol{\theta} \), and \( \odot \) represents the Hadamard (elementwise) product.  
This construction guarantees exact invertibility at machine precision and enables stable, frequency-wise control of the learned filters.

This parameterization ensures that the decoder can apply exact inverse filtering in Fourier space by elementwise division:
\[
\widehat{\mathbf{f}}' = \frac{\widehat{\mathbf{z}}}{\widehat{\mathbf{w}}}, \qquad
\mathbf{f}' = \mathcal{F}^{-1}(\widehat{\mathbf{f}}').
\]
Because convolution commutes with translation, i.e.,$ \mathcal{W}[\mathcal{T}_{\boldsymbol{a}} f] = \mathcal{T}_{\boldsymbol{a}}[\mathcal{W}f], \quad \forall a \in \Omega$, translation equivariance is naturally preserved at every layer. 

\subsection{Training Strategy}
\label{sec:training}

All models were trained under invertibility-preserving constraints using the Adam optimization algorithm.  
The training objective minimized the mean-squared reconstruction error between the input and the decoded signal, subject to implicit constraints ensuring bijectivity at every layer.  
Specifically, the spectral parameters \( \boldsymbol{\theta} \) of the convolutional filters (see Section~\ref{sec:filters}) were updated under the exponential parameterization \( \widehat{w}(\boldsymbol{k}) = e^{\theta(\boldsymbol{k})} \), guaranteeing nonvanishing Fourier coefficients throughout training.  
Monotonic activations were simultaneously regularized through their Lipschitz constants to preserve the conditions for invertibility described in Section~\ref{sec:activation}.

For one-dimensional problems, where the data dimensionality is moderate for efficient training, the training was performed in full-batch mode to ensure stable gradient propagation and exact enforcement of spectral normalization across all layers.  

For higher-dimensional or spatially extended problems, such as the two-dimensional toy dataset, mini-batch training was employed to balance convergence stability and computational efficiency.  
Mini-batches of size 64 were sampled with randomly shuffled phase realizations at each epoch, enabling the network to encounter a broad distribution of spatial translations and frequency combinations.  
The learning rate was initialized at \(10^{-3}\) and decayed geometrically by a factor of 0.9 every 20 epochs.  
Spectral normalization and monotonicity regularization were applied at each update step, ensuring that all invertibility guarantees derived in this section were maintained throughout training.

\section{Experiments}
\subsection{Toy problem: 1D Sinusoidal Mixture}
\label{sec:Toy1}

The first example investigated is a synthetic one-dimensional signal defined by
\[
f(x; \boldsymbol{\omega}) = \tanh\Bigl(\sin\bigl(x + \omega_1 + \cos(2x + \omega_2)\bigr)\Bigr)
\]
where \( \boldsymbol{\omega} = (\omega_1, \omega_2) \in \Theta=[0,2\pi)^2 \) denotes the phase of two nonlinearly interacting waveforms, and \( x \in [0,2\pi)\) is the independent variable. The dataset is generated by randomly sampling 100 different parameter sets \( \boldsymbol{\omega} \), and evaluating the corresponding signals on a uniform grid of 128 points in \( x \in [0, 2\pi) \).
This dataset is particularly well-suited for our study for several reasons. First, the underlying latent space is two-dimensional and periodic in both directions, naturally forming a torus topology. Recovering this toroidal structure is an essential challenge for any method aiming to uncover meaningful low-dimensional representations of the data. Second, the signal is highly nonlinear due to the nested sine and cosine functions and the outer hyperbolic tangent, leading to nontrivial interactions between frequency components.

\begin{figure*}
\centering
\makebox[\textwidth]{%
  \includegraphics[width=1.0\textwidth]{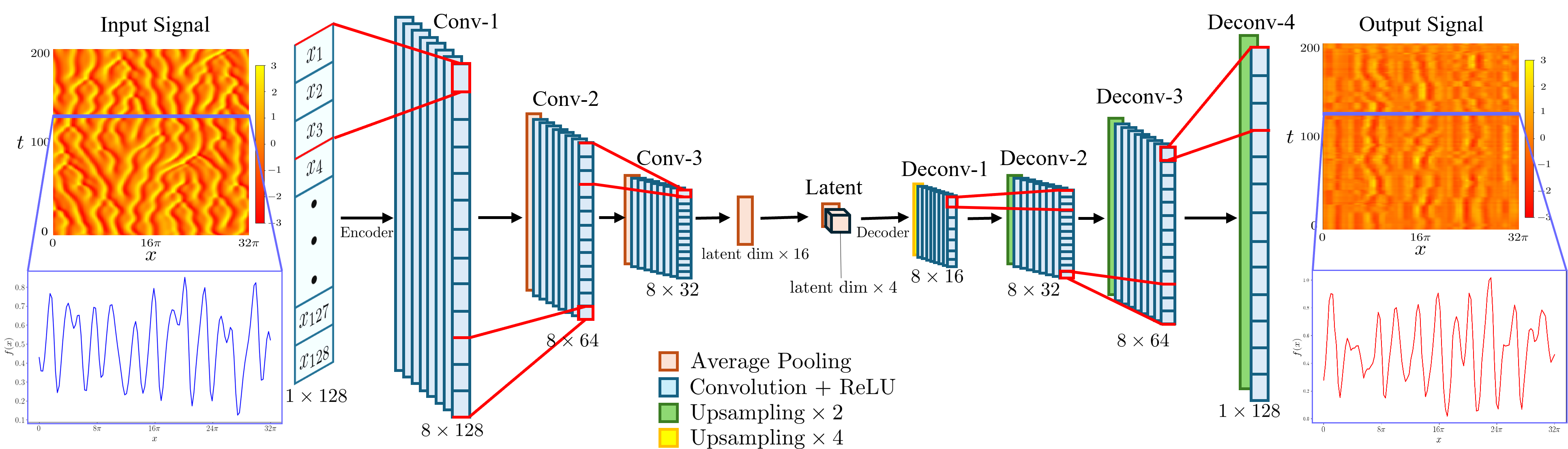}
}
\caption{\centering The architecture of the CNN models used in comparison in the one-dimensional K-S dataset.}
\label{fig:model_structures}
\end{figure*}
Traditional linear decomposition methods, which all result in the discrete Fourier transform (DFT), fail to capture the underlying structure of the data. In particular, the energy spectrum obtained from the FFT suggests that the second harmonic is entirely absent, despite the fact that a term with wave number 2 is present inside the composition: the nonlinear composition effectively cancels out the signal with wave number $2$. This highlights the limitations of linear methods in representing data governed by nonlinear and phase-dependent systems, and motivates the use of more expressive tools capable of resolving hidden geometric and topological features in function space.

The goal of this toy problem is to evaluate how different autoencoder architectures compress and reconstruct structured, nonlinear signals using a low-dimensional latent space. In all experiments, the latent dimension is 2, recognizing that the zero wavenumber component is consistently $0$ across the dataset.

We compare the FINE model with conventional CNN-based down-sampling and up-sampling autoencoders, consisting of a stack of 1D convolutions, max-pooling layers, and transposed convolutions, as shown in figure \ref{fig:model_structures}.

\begin{figure}[h]
\centering
    \centering
    \includegraphics[width=1.0\textwidth]{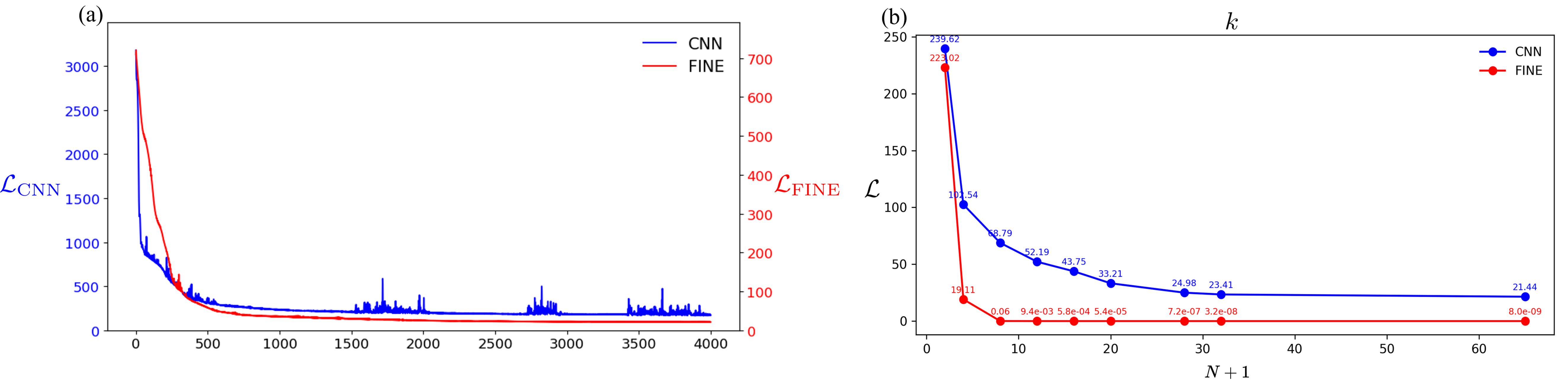}
\label{fig:toy_1_loss_comparison}
\caption{
Performance comparison for different latent space between convolutional neural network (CNN) and Fourier-Invertible neural encoder (FINE) models on Toy Problem. }
\label{fig:loss_comparison}
\end{figure}

\textbf{Loss Curve Comparison}
Figure~\ref{fig:loss_comparison}$(a)$ shows the training loss curve of the CNN autoencoder over 4000 epochs in the blue line. The training begins with a high loss of approximately 3000 and decreases rapidly during the initial phase (first 500 epochs), indicating that the model quickly captures coarse features of the signal. Subsequently, the loss continues to decrease gradually, reaching a drop of about 90\% at epoch 4000. To complement this analysis, we compare the final reconstruction losses of CNN and FINE models across varying latent dimensions, as illustrated in Figure~\ref{fig:loss_comparison}$(b)$. The FINE model achieves near-zero reconstruction loss with as few as 8 latent dimensions, and the loss remains at machine precision for all higher dimensions. In contrast, the CNN model requires significantly more latent capacity to reduce the loss, and even at dimension 64, it fails to match FINE’s performance. These results highlight FINE’s superior ability to extract and represent essential latent features efficiently, demonstrating its effectiveness in capturing the underlying structure of the signal with fewer parameters.

In contrast, the training loss curve of the FINE model is shown in the red line. The FINE model starts with a significantly lower initial loss and rapidly converges to below 100 within the first 300 epochs. The overall convergence is faster and smoother compared to the CNN model, with more than ten times the accuracy (1/10 of the final loss).

\textbf{Reconstruction Quality at Convergence}
To qualitatively assess the final reconstruction ability of the two models, we visualize a representative test signal reconstructed at the end of training (epoch 4000). 

\begin{figure}
\centering
\makebox[0.9\textwidth]{%
  \includegraphics[width=0.9\textwidth]{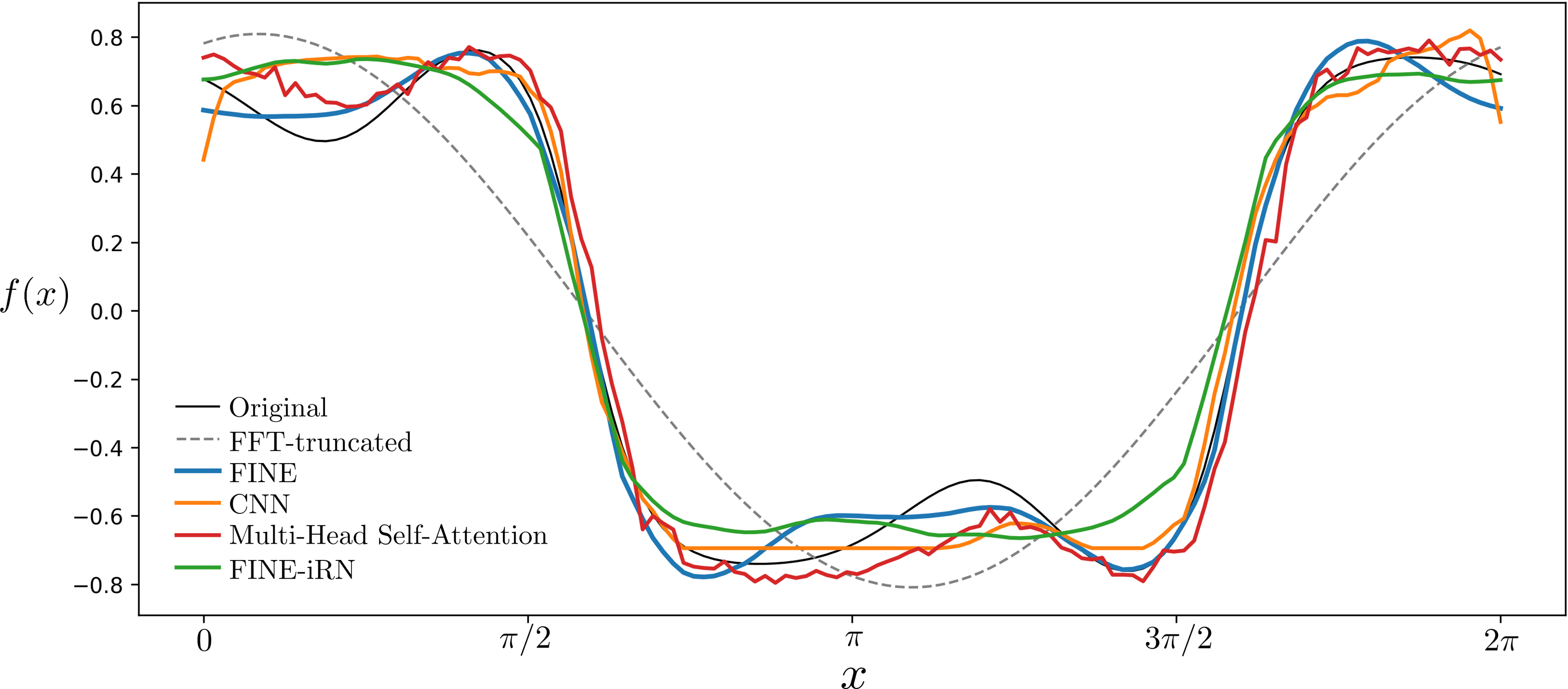}
}
\caption{Reconstruction results on the Toy Problem 1, using a latent dimension of $N=2$.}
\label{fig:Toy_reconstruction}
\end{figure}

Figure~\ref{fig:Toy_reconstruction}  Reconstruction results on Toy Problem 1 using a 2-dimensional latent representation after 4000 training epochs.
The original signal $f(x)$ (black) is compared against reconstructions from six models: FFT-truncated (gray), CNN (orange), FINE (blue), Self-Attention (red), and FINE-IRN (green). The CNN-based autoencoder exhibits underfitting near regions of rapid variation, particularly around local extrema, indicating limited capacity to capture global structure in low-dimensional latent space.
In contrast, the FINE model reconstructs the signal with significantly higher fidelity. It accurately preserves the phase, amplitude, and harmonic content, closely matching the original waveform even in high-frequency regions. Despite having fewer parameters than the CNN, FINE benefits from its invertible architecture and Fourier-aware latent structure. This comparison underscores the limitations of conventional CNN autoencoders for structured signals and highlights the representational efficiency of FINE.

As an additional baseline, we also evaluated an FINE-IRN(iResNet-based autoencoder). FINE-IRN achieves a notably improved reconstruction over the CNN, accurately capturing local extrema and waveform smoothness. Although slightly less precise than the FINE model in high-frequency regions, it demonstrates that invertible architectures are effective in low-dimensional latent spaces even without explicit spectral structure. This further reinforces the advantage of the FINE design, which combines invertibility with Fourier-domain awareness to reach superior reconstruction fidelity.

\begin{table} 
\centering
\begin{tabular}{l|c|c|c}
\hline
\textbf{Metric} & \textbf{CNN} & \textbf{FINE} & \textbf{Gain (FINE/CNN)} \\
\hline
Parameters & 812 & 109 & 0.13$\times$ \\
CPU Time& 24.70 s& 51.16 s& 2.07$\times$ \\
MSE & 173.47 & 14.20 & 0.08$\times$ \\
\hline
\end{tabular}
\vspace{10 pt}
\caption{Comparison of performance metrics between CNN and FINE on Toy Problem 1.}
\vspace{-10 pt}
\label{tab:toy_1_cnn_fine_comparison}
\end{table}

Table~\ref{tab:toy_1_cnn_fine_comparison} summarizes the performance of the proposed FINE model compared to a conventional CNN on Toy Problem 1. While FINE requires longer training time (approximately 2× in CPU time), this is an expected outcome given the model's architectural constraints and initialization strategy, as discussed in previous sections. Despite the increased training cost, FINE achieves substantially improved accuracy, with the mean squared error (MSE) reduced to approximately 8\% of that of CNN, and it does so with only about 13\% of the trainable parameters. These results reflect a trade-off between computational efficiency and model interpretability: FINE imposes structural biases that better capture the underlying symmetries of the data, making it particularly suitable for applications where understanding the learned representation is as important as predictive performance.

\subsection{One-dimensional turbulence in Kuramoto–Sivashinsky equation}
\label{sec:ks}
The Kuramoto--Sivashinsky (K--S) equation,
\[
\frac{\partial u}{\partial t} + u \frac{\partial u}{\partial x} + \frac{\partial^2 u}{\partial x^2} + \nu \frac{\partial^4 u}{\partial x^4} = 0,
\]
is a canonical one-dimensional nonlinear partial differential equation used to model instabilities in laminar flame fronts, reaction–diffusion systems, and thin film flows.  
It exhibits chaotic dynamics characterized by the coexistence of long-wave instabilities and short-wave dissipation, making it an established benchmark for studying nonlinear model reduction.  

\begin{wrapfigure}{r}{0.43\textwidth}
    \vspace{-10pt}
    \centering
    \includegraphics[width=0.42\textwidth]{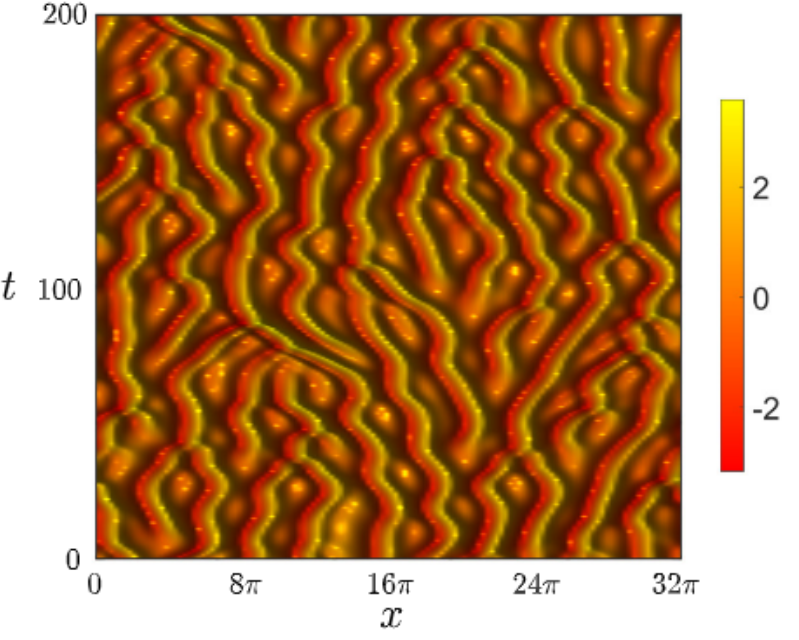}
    \vspace{-5pt}
    \caption{Sample evolution of the K--S equation solution.}
    \label{fig:ks_solution_wrap}
\end{wrapfigure}

We solved the K--S equation on a periodic domain with 128 uniformly spaced spatial points and collected 1000 temporal snapshots from a trajectory of total duration \(10^4\).  
A segment of this solution is shown in Figure~\ref{fig:ks_solution_wrap}.  
Each snapshot represents a different realization of a translation-equivariant turbulent field, \( f_t(x) = u(x,t) \), containing rich multiscale content and strong nonlinear phase interactions, providing a challenging benchmark for nonlinear model reduction.

The goal of this experiment is to assess whether the proposed FINE architecture can reconstruct such chaotic dynamics more accurately and efficiently than conventional convolutional autoencoders.  
Table~\ref{tab:toy_2_cnn_fine_comparison} summarizes the quantitative results for a latent dimension of ten.  
Despite having nearly an order of magnitude fewer parameters (174 versus 1,496), FINE achieves a mean-squared reconstruction error of 550.3, corresponding to roughly one-fifth of the error obtained with the CNN model.  
The improved accuracy comes at the cost of longer training time, approximately an order of magnitude higher on both CPU and GPU, reflecting the additional cost of enforcing invertibility and spectral normalization.  
Nevertheless, the dramatic reduction in error and parameter count highlights the efficiency of FINE’s spectral representation.

\begin{figure}[h]
\includegraphics[width=1.0\textwidth]{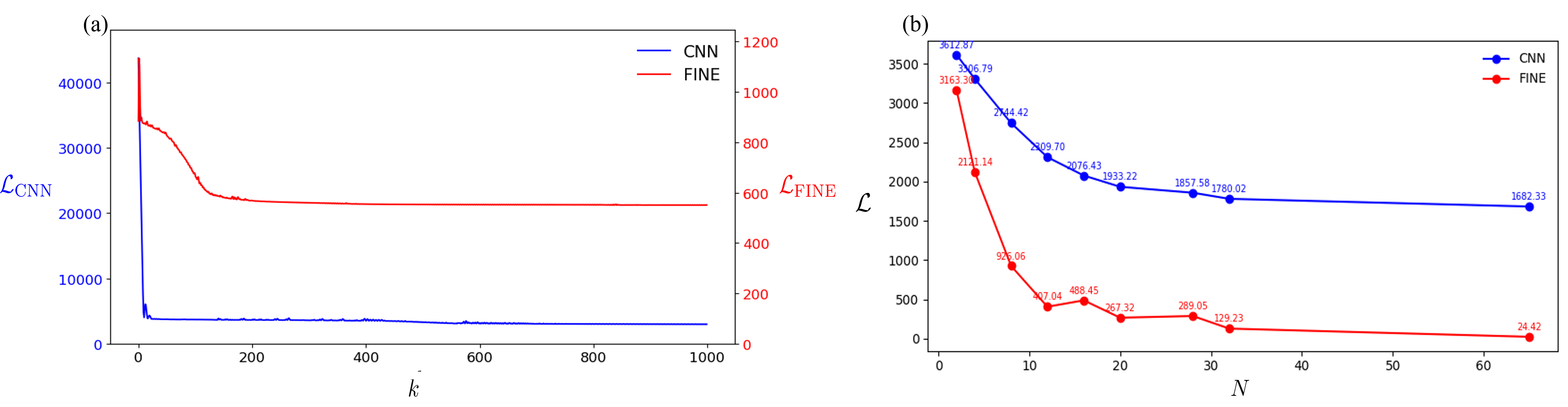}
\caption{(a) Training loss trajectories $\mathcal{L}_{\mathrm{CNN}}$ and $\mathcal{L}_{\mathrm{FINE}}$ over epochs k on the solution of K-S equation. (b) Performance comparison between a convolutional neural network (CNN) and the FINE model on the solution of K-S equation.}
\label{fig:ks_loss_comparison}
\end{figure}

\begin{table} 
\centering
\begin{tabular}{l|c|c|c}
\hline
\textbf{Metric} & \textbf{CNN} & \textbf{FINE} & \textbf{Gain (FINE/CNN)} \\
\hline
Parameters & 1496 & 174 & 0.12× \\
CPU Time & 7.23s & 80.72s & 11.16× \\
GPU Time & 5.09s & 56.86s & 11.17× \\
MSE & 2669.23 & 550.30 & 0.21× \\
\hline
\end{tabular}
\vspace{10 pt}
\caption{Comparison of performance metrics between the CNN and FINE models on K-S Problem.}
\label{tab:toy_2_cnn_fine_comparison}
\end{table}
The convergence behavior of the two models is shown in Figure~\ref{fig:ks_loss_comparison}.
The CNN autoencoder begins with a high loss exceeding 40,000 and converges slowly, exhibiting large oscillations indicative of unstable gradient propagation.  
In contrast, FINE starts from a much smaller initial loss of approximately 1150 and converges smoothly to about 556 after 1,000 epochs.  
The steadier decay reflects the stabilizing effect of invertible activations and the Fourier-based representation, which together constrain the learning dynamics to a well-conditioned manifold of reversible transformations.  
The comparison of reconstruction error versus latent dimension further demonstrates that FINE reaches a low-loss regime with substantially smaller latent spaces, confirming that its Fourier truncation bottleneck captures the most energetically significant modes efficiently.

\begin{figure}[h]
\centering    \includegraphics[width=0.9\textwidth]{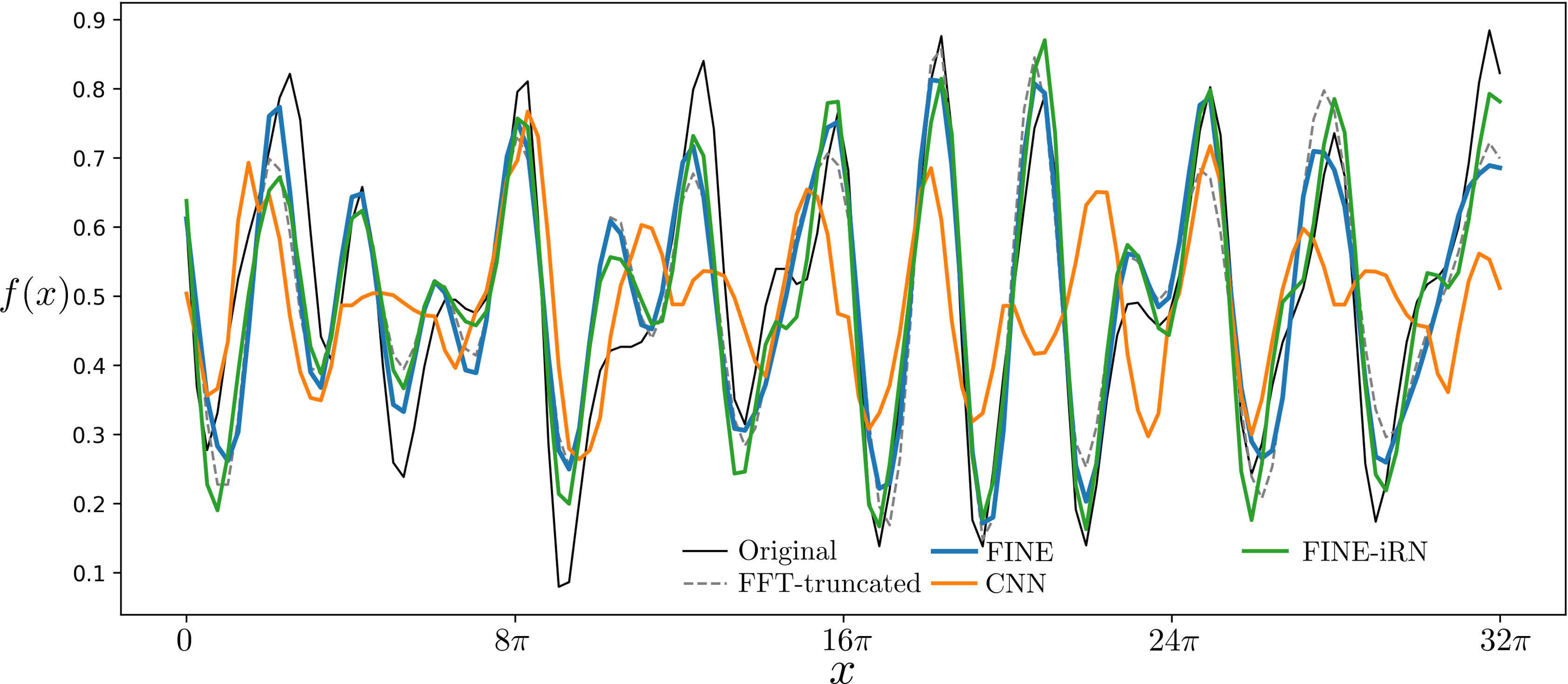}
\caption{Reconstruction result on the 1D Kuramoto–Sivashinsky dataset.}
\label{fig:ks_reconstruction}
\end{figure}
We evaluate the reconstruction performance of both models at epoch 1000 using a representative K-S snapshot. Each figure overlays the ground truth signal, its Fourier-truncated baseline, and the reconstructed output.
Figure~\ref{fig:ks_reconstruction} shows the reconstruction results of the CNN-based autoencoder, the FINE model, and the FINE-IRN on a signal from the statistically stationary solution of Kuramoto–Sivashinsky (KS) equation. The CNN reconstruction captures the overall trend but significantly smooths the amplitude envelope, particularly in high-frequency regions. Sharp peaks are often damped or misaligned, indicating the difficulty of CNN to retain critical turbulent modes under severe dimensional compression. The FINE reconstruction exhibits precise alignment with the original waveform. It successfully recovers both dominant and localized features, closely tracking rapid transitions and preserving high-frequency components. Compared to the Fourier-truncated baseline, FINE offers superior sharpness and phase fidelity, demonstrating the effectiveness of invertible spectral filtering and monotonic nonlinearities in structured signal modeling.
In comparison, the FINE-IRN model achieves comparable reconstruction quality, showing that invertible networks with elementwise updates can effectively model nonlinear dynamics. Nevertheless, FINE remains the core contribution: its structured spectral filtering and monotonic activations yield sharper reconstructions and better capture translation symmetry. These results affirm FINE’s extensibility and effectiveness for learning low-dimensional representations of turbulent signals.

\begin{figure}[h]
\centering
\includegraphics[width=0.8\textwidth]{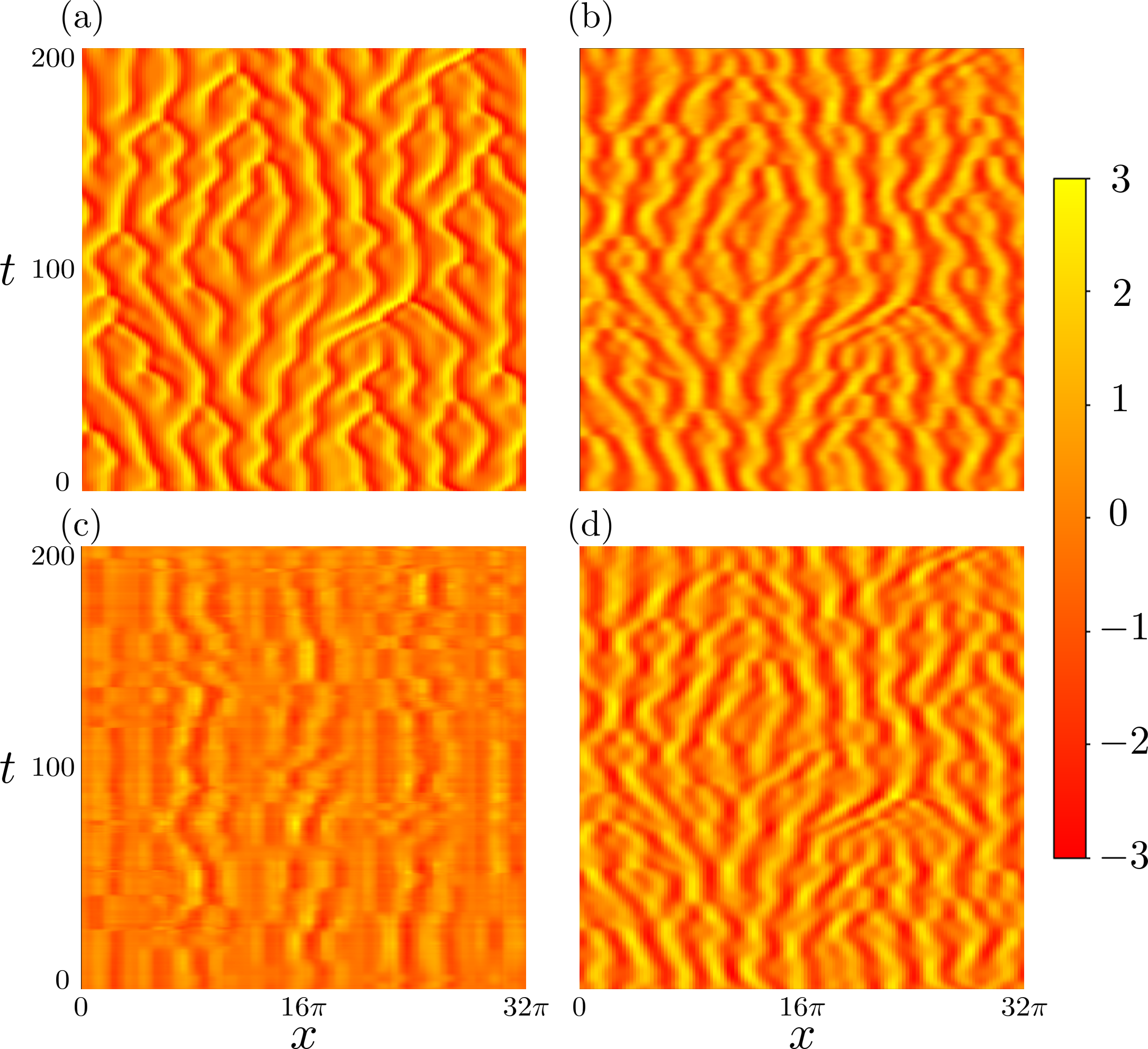}
\caption{Two-dimensional contour plots of the reconstructed solutions of the Kuramoto-Sivashinsky (K-S) equation obtained using different models. Each panel compares the reconstructed field with the reference solution, illustrating differences in spatial accuracy and phase fidelity across models. Panel (a) shows the ground truth; (b) presents the FINE reconstruction; (c) displays the CNN reconstruction; and (d) depicts the FINE-IRN reconstruction.}
\label{fig:KS_2dreconstruction}
\end{figure}

To visualize the spatiotemporal coherence of the reconstructions, Figure~\ref{fig:KS_2dreconstruction} presents contour plots of the space–time evolution for different models.  
The CNN reconstruction shows blurred ridges and temporal phase shifts, leading to visible misalignment of traveling-wave structures.  
FINE, on the other hand, preserves the sharpness of the spatiotemporal ridges and reproduces the correct phase velocity across the entire domain, indicating consistent translation-equivariant encoding in its latent representation.  
This visual agreement supports the quantitative findings: FINE maintains the integrity of the turbulent dynamics even under strong dimensional reduction.

\begin{figure*}
\centering
\includegraphics[width=\textwidth]{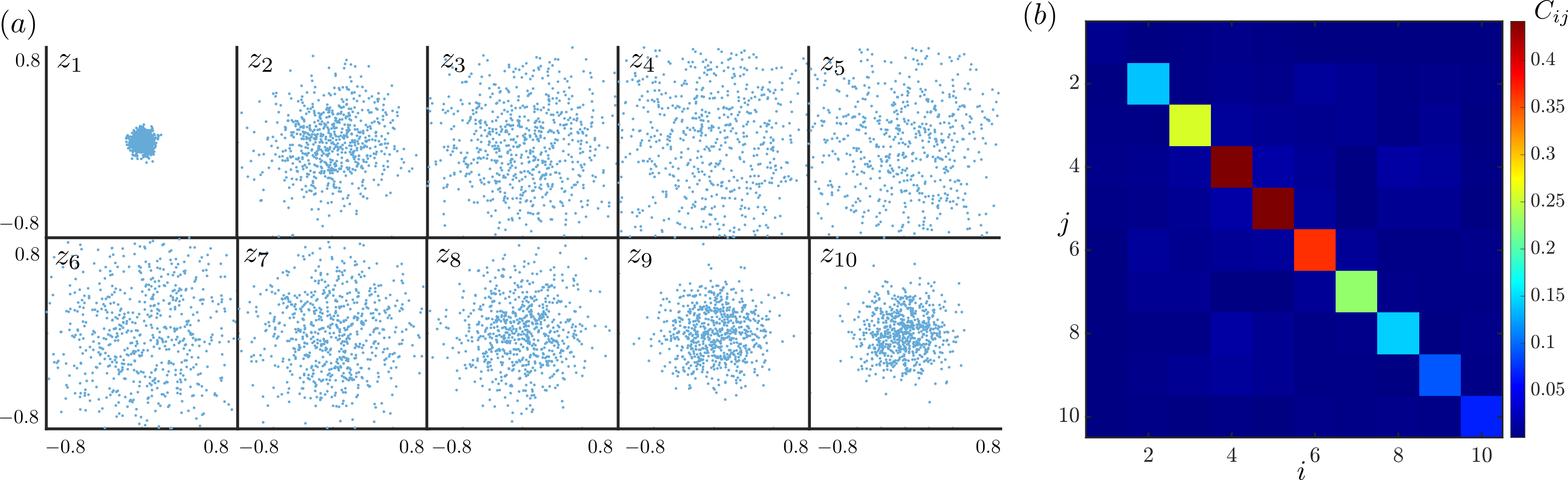}
\caption{(a) Visualization of the FINE model’s latent space distribution on the statistically stationary solution of K-S equation. (b) The covariance, $C_{ij} = |\langle  z_i^* z_j  \rangle|$ in the latent space of this K-S dataset.}
\label{fig:latent_decode_reconstruction}
\end{figure*}

Figure~\ref{fig:latent_decode_reconstruction} illustrates the latent space distribution learned by the FINE model for the one-dimensional Kuramoto-Sivashinsky (K-S) problem. Each subplot corresponds to a different complex-valued latent variable: the left plot shows the distribution of \( z_1 \) and the right plot shows \( z_2 \), on a complex plane. Notably, both latent variables exhibit a Gaussian distribution in the latent space and remain uncorrelated. This behavior suggests a similarity with the $\beta$-Variational Autoencoder (\citep{solera2024beta}), in that the latent dimensions are independent of each other. 
When decomposed into Fourier modes in the latent space, the model exhibits approximately independent latent coordinates, 
indicating an underlying generative structure--a direction we leave for future investigation.

\subsection{Two-dimensional synthetic field with separable trigonometric structure}
\label{sec:toy2d}

The second benchmark is a synthetic two-dimensional field designed to test the ability of FINE to recover multi-dimensional, translation-equivariant structure from nonlinear mixtures.  
The ground-truth signal is defined as
\begin{equation}
\begin{aligned}
f(x,y;\boldsymbol{\omega})
=&\operatorname{arctanh}\!\Big(
\sin(x+\omega_1)\sin(y+\omega_2)
\\
&+\cos(2x+\omega_3)\cos(2y+\omega_4)
\Big),
\end{aligned}
\end{equation}
where the phase parameters
\(\boldsymbol{\omega}=(\omega_1,\omega_2,\omega_3,\omega_4)\in[0,2\pi)^4\)
control independent shifts in both directions, and
\((x,y)\in[0,2\pi)\times[0,2\pi)\)
are spatial coordinates on a periodic domain.  
Each realization \(f_{\boldsymbol{\omega}}\) is evaluated on a uniform grid of size \(H\times W\), forming the discrete dataset
\(\mathcal{D}=\{f_{\boldsymbol{\omega}_m}\}_{m=1}^{M}\).

This dataset is ideal for assessing translation-equivariant representation learning.  
Before the nonlinear transformation, the field consists of only two separable harmonic components with wavenumbers \((1,1)\) and \((2,2)\); the outer \(\operatorname{arctanh}\) introduces nonlinear coupling among these modes, enriching the spectrum and generating higher-order harmonics.  
The intrinsic latent space is thus four-dimensional and periodic, topologically a product torus \(\mathbb{T}^{4}\).  
Recovering this low-dimensional manifold from high-dimensional grid samples constitutes a clear test of whether a model can learn phase-dependent structure rather than simply compress amplitude patterns.  
Linear decompositions such as POD or DFT distribute the energy over multiple oblique wavenumber pairs \((m,n)\) and cannot retrieve the generating phase coordinates.  
For example,
\(\sin x\,\sin y=\tfrac12[\cos(x-y)-\cos(x+y)]\)  
and
\(\cos 2x\,\cos 2y=\tfrac12[\cos(2(x-y))+\cos(2(x+y))]\);
the nonlinear composition further mixes these diagonal components, producing broadband spectra that obscure the latent toroidal coordinates.

We compared FINE against two alternative architectures: a conventional convolutional autoencoder (CNN) and a self-attention autoencoder. The CNN consists of four stride-2 convolutional downsampling blocks with channel progression \(1 \rightarrow 32 \rightarrow 64 \rightarrow 128 \rightarrow 256\), followed by a \(\frac{H}{16} \times \frac{W}{16}\) bottleneck projected into a latent vector. The decoder mirrors this structure using transposed-convolution upsampling layers. This design provides sufficient hierarchical capacity while maintaining a parameter count comparable to the attention model, though the downsampling operations inevitably reduce spectral resolution at intermediate layers. For the self-attention autoencoder, a direct pixel-level attention model, in which each point on the grid is treated as a token, was computationally infeasible for realistic size grids due to the scaling \(\mathcal{O}((HW)^2)\) of the attention matrix.
Instead, we implemented a patch-based variant with localized windowed tokens.  
While this reduces the computational cost, self-attention remains \emph{permutation-equivariant} rather than \emph{translation-equivariant}; its responses depend only on pairwise correlations between tokens and not on their absolute or cyclic positions.

Therefore, the attention model tends to over-parametrize the functions for a translation-equivariant dataset, resulting in potential waste of training power. 
In our experiments, this over-parametrization leads to increased difficulty in converging and visible spatial artifacts compared with FINE.

\begin{table} 
\vspace{-10pt}
\centering
\begin{tabular}{l|c|c|c}
\hline
 \textbf{Metric} & \textbf{CNN} & \textbf{FINE} & \textbf{Gain (FINE/CNN)} \\
\hline
Parameters           & 950{,}758 & 16{,}684 & 0.0175× \\
CPU Time& 45.66s     & 53.82s  & 1.18×   \\
Loss (MSE)           & 3075.23   & 6.35   & 0.00207× \\
\hline
\end{tabular}
\vspace{10 pt}
\caption{Performance comparison on the two-dimensional synthetic field.}
\vspace{-10 pt}
\label{tab:toy3_cnn_fine_comparison}
\end{table}
Table~\ref{tab:toy3_cnn_fine_comparison} summarizes the quantitative comparison between CNN and FINE on this two-dimensional problem.  
Although FINE contains nearly two orders of magnitude fewer parameters (16{,}684 versus 950{,}758), it achieves over three orders of magnitude smaller reconstruction error (MSE 6.35 compared with 3075.23) while requiring comparable CPU training time.  
This remarkable efficiency stems from FINE’s spectral parameterization, which enforces invertibility and translation equivariance by design, allowing the network to isolate the true low-dimensional phase manifold with minimal redundancy.

\begin{figure}[h]
    \centering
    \includegraphics[width=0.9\textwidth]{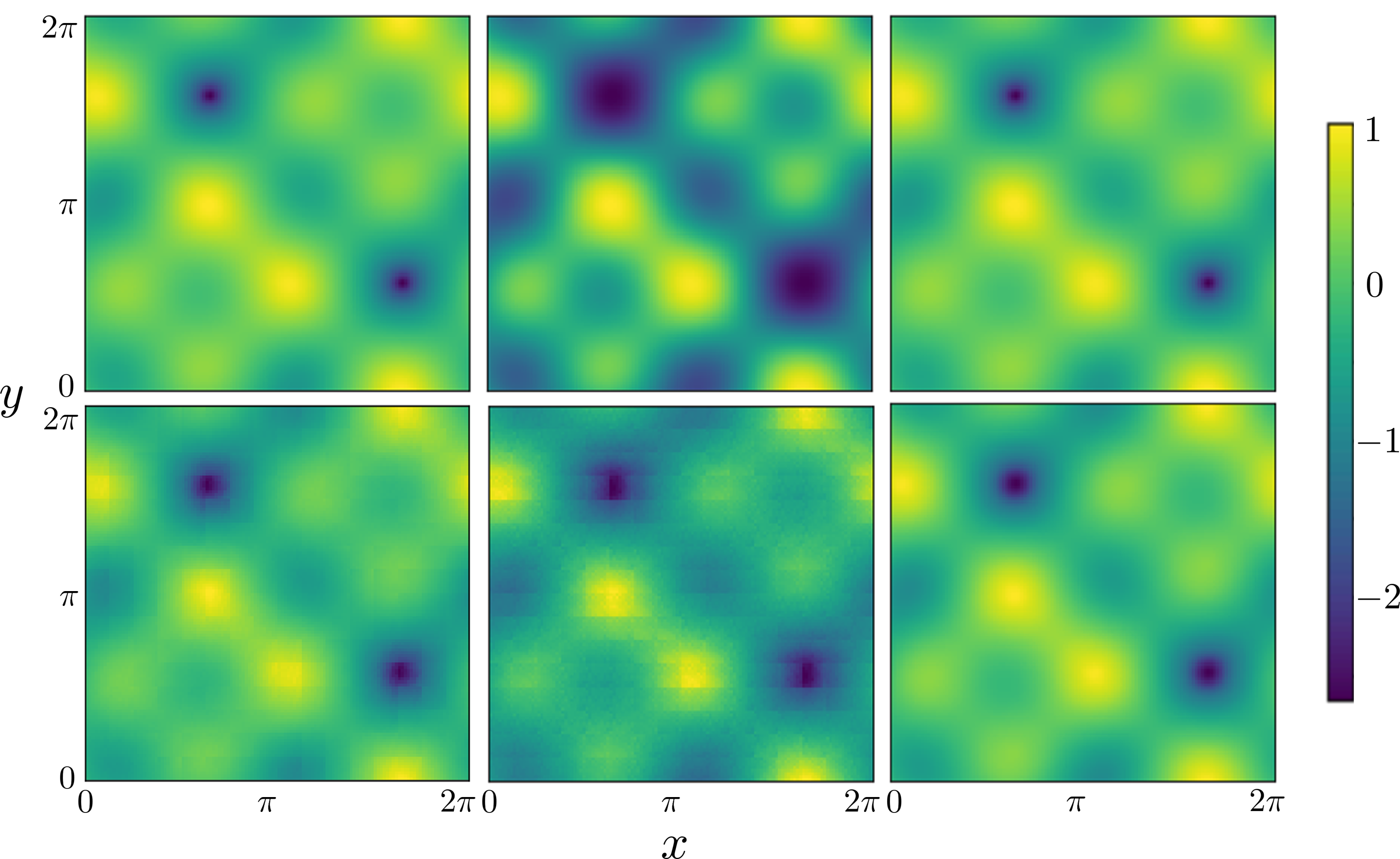}
    \caption{
    Comparison with the ground truth on the 2D synthetic dataset. 
    Top row: original field, reconstructions using POD (FFT truncation), and FINE. 
    Bottom row: reconstructions using Attention, CNN, and FINE-IRN. 
    Colored contours indicate the function value \( f \).
    }
    \label{fig:p3_reconstruction}
\end{figure}

Figure~\ref{fig:p3_reconstruction} compares reconstructions obtained by the different models for a representative sample at latent dimension \(d=5\) (one bias term plus four latent coordinates corresponding to the underlying phases).  
FINE nearly reproduces the true field exactly, accurately matching both the amplitude and the fine-scale phase variations.  
The Fourier-truncated baseline retains only the dominant low-frequency structures but fails to reconstruct high-frequency harmonics introduced by the nonlinear transformation.  
The CNN model produces visibly noisy outputs with loss of separability and cross-directional interference, whereas the attention-based autoencoder displays block-wise artifacts associated with its patch partitioning and the absence of strict translation equivariance.  
An invertible residual network (FINE-IRN2D) yields smoother reconstructions than the CNN but still lags behind FINE, which uniquely integrates spectral filtering with invertibility and monotonic activations.  
Visually, the FINE output is indistinguishable from the ground truth, confirming that the model captures both global phase relations and local nonlinear interactions.

\begin{figure}[h]
\centering
\includegraphics[width=1.0\textwidth]{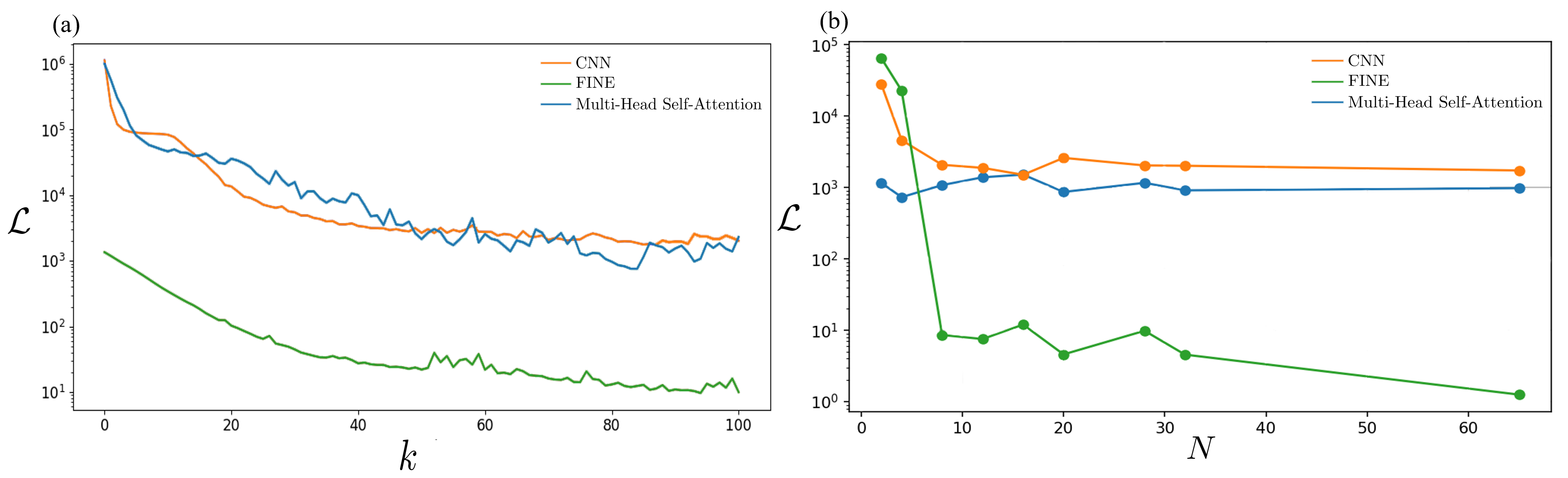}
\caption{Training dynamics and dimensional-dependence of reconstruction loss for the two-dimensional benchmark. (a)Training loss evolution.(b)Reconstruction loss versus latent dimension.}
\label{fig:loss_3_comparison}
\end{figure}

The training behavior of each model is reported in Figure~\ref{fig:loss_3_comparison}.  
FINE converges rapidly to a loss nearly three orders of magnitude lower than both CNN and attention variants, indicating that it successfully identifies the four intrinsic degrees of freedom.  
For latent dimensions \(d<5\), the loss remains high because the latent space cannot fully represent the four phase parameters.  
Once \(d=5\) is reached, the reconstruction error drops sharply and saturates, marking the point where the network has captured all effective degrees of freedom.  
Beyond this dimension, additional latent variables offer negligible improvement, reflecting the compactness of the learned toroidal representation.  
In contrast, CNN and attention autoencoders show only marginal loss reduction with increasing \(d\), revealing that their representations are not aligned with the underlying physical or spectral structure.  
These results highlight the advantage of incorporating explicit Fourier-domain structure: FINE approaches the theoretical representational limit using a small latent space, whereas conventional architectures require far larger capacity to approximate the same mappings.

In summary, this two-dimensional example demonstrates that FINE effectively generalizes its invertible, translation-equivariant design to multivariate fields.  
By parameterizing filters in the spectral domain and constraining nonlinear activations to be monotonic and invertible, FINE captures the underlying toroidal phase structure with unprecedented efficiency, outperforming convolutional and attention-based architectures both quantitatively and qualitatively.

\subsection{Two-dimensional isotropic turbulence}
\label{sec:2diso}
To further assess the capability of the proposed model in realistic multi-scale physics, we evaluate reconstruction performance on two-dimensional decaying isotropic turbulence, a canonical benchmark for nonlinear energy cascade and vortex stretching dynamics. The flow is generated by solving the incompressible Navier-Stokes equations on a doubly periodic domain \( [0, 2\pi]^2 \),
\[
\frac{\partial \mathbf{u}}{\partial t} + (\mathbf{u} \cdot \nabla)\mathbf{u} = -\nabla p + \frac{1}{Re} \nabla^2 \mathbf{u}, \quad \nabla \cdot \mathbf{u} = 0,
\]
with \( Re = 1000 \). The initial field is synthesized from a prescribed energy spectrum, with a grid size of \( 128 \times 128 \).

The dataset is collected for the vorticity snapshots at time $t = 0.1$. Each snapshot contains rich multi-scale structures, which is the main challenge of dimension reduction in such setups.

\paragraph{CNN Autoencoder Architecture and Parameter Settings}
The CNN baseline used in the 2D turbulence experiment is a symmetric convolutional autoencoder composed of four stages of stride-2 downsampling, followed by four corresponding transposed-convolution upsampling stages. The encoder progressively maps the input vorticity field of size \(128 \times 128\) to a feature map of size \(256 \times 8 \times 8\), which is then collapsed into a 20-dimensional latent vector via a final \(8 \times 8\) convolution. The decoder mirrors this hierarchy, expanding the latent vector back to the full spatial resolution.

\begin{figure}[h]
\includegraphics[width=1.0\textwidth]{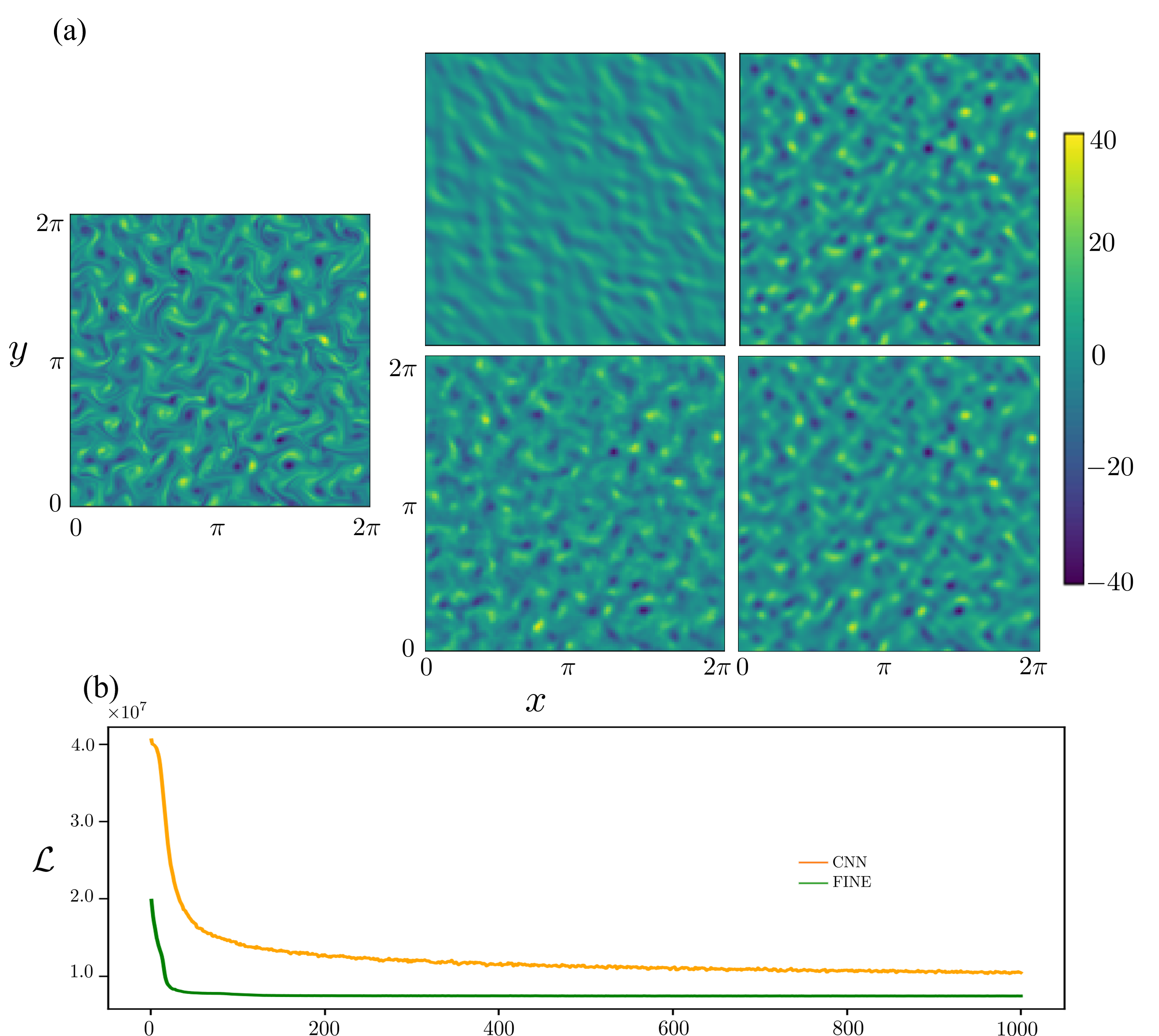}
    \caption{(a) Comparison for the reconstructed 2D incompressible decaying turbulence, in terms of vorticity field. Left: Original. Top row: FFT, FINE. Bottom row: CNN, FINE-IRN. (b) Convergence history of different models during training.}
    \label{fig:p4_reconstruction}
\end{figure}

We use a common latent dimension of 20 across all models. Consistent with the behavior observed in earlier synthetic benchmarks, the FFT baseline recovers only the dominant large-scale vortices and fails to reproduce small-scale enstrophy. The CNN autoencoder provides smoother reconstructions but oversuppresses inertial-range structures due to downsampling and strided convolutions. In contrast, FINE and FINE-iRN achieve the closest agreement with the ground truth, accurately reproducing coherent vortices and delicate trailing filaments without introducing ringing, aliasing, or blockwise distortions. These results reinforce the architectural advantages of FINE.

\begin{table}[H]
\centering
\begin{tabular}{l|c|c|c}
\hline
\textbf{Metric} & \textbf{CNN2D} & \textbf{FINE} & \textbf{Gain (FINE/CNN)} \\
\hline
Parameters           & 1{,}442{,}293 & 33{,}346 & $0.023 \times$ \\
CPU Time             & 24.8h     & 25.6h & $1.03 \times$ \\
MSE                  & $1.2\times 10^7$  & $5.0\times 10^6$ & $0.42 \times$ \\
\bottomrule
\end{tabular}
\vspace{10 pt}
\caption{Computational cost, model efficiency, and relative gain of FINE over CNN.}
\vspace{-10 pt}
\label{tab:params_timing}
\end{table}

The corresponding convergence behavior in Figure~\ref{fig:p4_reconstruction} further highlights the difference among architectures. The CNN and attention models display slow, saturated convergence, while FINE achieves rapid, smooth loss decay and reaches an error level nearly two orders of magnitude smaller. The stability and monotonicity of FINE’s training curve reflect the well-conditioned nature of its invertible spectral operators and monotonic nonlinearities. FINE-IRN reaches similar final reconstruction accuracy as FINE, but its training curve is noisy due to fixed-point inversion and Lipschitz-constrained updates. Since the convergence history is not directly comparable across models and adds no new insight beyond the visual results, we omit it for clarity.

In terms of computational efficiency, FINE-2D is remarkably compact. It uses approximately \(10\times\) fewer parameters than the attention autoencoder and \(43\times\) fewer than the CNN autoencoder—yet achieves comparable training time across all models. This efficiency stems from its lightweight spectral blocks and Fourier-space operations. These findings underscore a central conclusion of this study: high-fidelity, physics-consistent reconstruction of turbulent flows does not require large, heavily parameterized architectures. Instead, the inductive biases in FINE enable it to outperform larger models while remaining significantly more parameter-efficient.

\section{Conclusion and Future Work}
\label{sec:conclusion}

This work introduced the \emph{Fourier–Invertible Neural Encoder (FINE)}, a new autoencoder framework that integrates invertible neural operators with a Fourier-domain bottleneck to achieve compact, interpretable, and symmetry-preserving representations of physical data.  
By construction, FINE is translation-equivariant: every spatial shift of the input signal corresponds to an equivalent shift in the reconstruction, ensuring that the learned latent space captures intrinsic dynamics rather than positional artifacts.  
Unlike conventional convolutional autoencoders, which rely on irreversible downsampling and fixed nonlinearities, FINE is composed entirely of invertible layers, namely monotonic activations and invertible convolutions, making all transformations bijective except for the explicit Fourier truncation step used for dimensional reduction.

Two alternative formulations for invertible activations were examined: a smooth monotonic ReLU approximation that guarantees differentiability and analytical invertibility, and a residual-type activation based on the Invertible ResNet (I-ResNet) architecture, which achieves invertibility through a Lipschitz-controlled perturbation of the identity mapping.  
Both approaches were shown to maintain stability during training and to preserve translation equivariance when applied pointwise to spatial coordinates.  
Together with the exponentially parameterized spectral filters, these activations ensure that FINE maintains exact information flow through the encoder–decoder pair and isolates information loss solely to the interpretable spectral truncation stage.

The proposed architecture was validated across multiple benchmarks of increasing complexity.  
For one-dimensional nonlinear wave interactions, FINE demonstrated the ability to recover the underlying toroidal phase structure with near-zero reconstruction loss.  
In the chaotic Kuramoto–Sivashinsky (K–S) system, FINE achieved up to an order of magnitude lower mean-squared reconstruction error compared with a convolutional autoencoder, while using only a fraction of the parameters.  
For the two-dimensional separable trigonometric benchmark, FINE generalized its symmetry-preserving design to multivariate fields, recovering the underlying four-dimensional latent torus with over three orders of magnitude improvement in accuracy relative to the CNN baseline.  

Beyond its empirical performance, FINE provides a conceptual bridge between several active research directions in scientific machine learning.  
It connects invertible neural networks, group-equivariant architectures, and operator-learning frameworks under a unified mathematical interpretation: dimensional reduction as a reversible transformation followed by an explicit spectral projection.  
This perspective offers a blueprint for building models that obey symmetries and reversibility constraints, which are properties fundamental to many physical or dynamical systems.  
The invertible and symmetry-preserving design of FINE also makes it a promising foundation for applications in data assimilation, reduced-order modeling, and generative reconstruction of physical fields from sparse or noisy observations.

Several potential paths for future research emerge from this study.  
First, a formal mathematical analysis of the convergence, stability, and expressivity of FINE is needed.  
While invertibility guarantees information preservation, a deeper understanding of how spectral truncation interacts with nonlinear invertible mappings would clarify error bounds and approximation rates.  
Second, scaling FINE to higher-dimensional and time-dependent problems such as two- and three-dimensional turbulent flows is needed.  
Integration with physics-informed constraints, including energy conservation and boundary conditions, could further enhance generalization and robustness.  
Finally, exploring hybrid formulations that combine FINE with operator-learning frameworks such as Fourier Neural Operators or Koopman autoencoders could yield architectures capable of both learning reduced representations and forecasting temporal evolution within a single, symmetry-perseving network.

\section*{Acknowledgements}
We gratefully acknowledge the support of the National Science Foundation under Grant No. NSF-2431610. 

\addcontentsline{toc}{section}{Acknowledgements}

\bibliographystyle{unsrtnat}
\bibliography{references}  

@String(ICLR = {Int. Conf. Learn. Represent.})

@String(ICLR  = {ICLR})

@inproceedings{satorras2021n,
  title={E (n) equivariant graph neural networks},
  author={Satorras, V{\i}ctor Garcia and Hoogeboom, Emiel and Welling, Max},
  booktitle={International conference on machine learning},
  pages={9323--9332},
  year={2021},
  organization={PMLR}
}

@inproceedings{cohen2019gauge,
  title={Gauge equivariant convolutional networks and the icosahedral CNN},
  author={Cohen, Taco and Weiler, Maurice and Kicanaoglu, Berkay and Welling, Max},
  booktitle={International conference on Machine learning},
  pages={1321--1330},
  year={2019},
  organization={PMLR}
}

@article{towne2018spectral,
  title={Spectral proper orthogonal decomposition and its relationship to dynamic mode decomposition and resolvent analysis},
  author={Towne, Aaron and Schmidt, Oliver T and Colonius, Tim},
  journal={Journal of Fluid Mechanics},
  volume={847},
  pages={821--867},
  year={2018},
  publisher={Cambridge University Press}
}

@article{hinton2006reducing,
  title={Reducing the dimensionality of data with neural networks},
  author={Hinton, Geoffrey E and Salakhutdinov, Ruslan R},
  journal={science},
  volume={313},
  number={5786},
  pages={504--507},
  year={2006},
  publisher={American Association for the Advancement of Science}
}

@article{fukami2019super,
  title={Super-resolution reconstruction of turbulent flows with machine learning},
  author={Fukami, Kai and Fukagata, Koji and Taira, Kunihiko},
  journal={Journal of Fluid Mechanics},
  volume={870},
  pages={106--120},
  year={2019},
  publisher={Cambridge University Press}
}

@article{motoori2021hierarchy,
  title={Hierarchy of coherent structures and real-space energy transfer in turbulent channel flow},
  author={Motoori, Yutaro and Goto, Susumu},
  journal={Journal of Fluid Mechanics},
  volume={911},
  pages={A27},
  year={2021},
  publisher={Cambridge University Press}
}

@article{park2025coherent,
  title={The coherent structure of the energy cascade in isotropic turbulence},
  author={Park, Danah and Lozano-Dur{\'a}n, Adri{\'a}n},
  journal={Scientific Reports},
  volume={15},
  number={1},
  pages={14},
  year={2025},
  publisher={Nature Publishing Group UK London}
}

@article{yu2021arbitrary,
  title={Arbitrary-Depth Universal Approximation Theorems for Operator Neural Networks},
  author={Annan Yu and Chloe Becquey and Diana Halikias and Matthew Esmaili Mallory and Alex Townsend},
  journal={ArXiv},
  year={2021},
  volume={abs/2109.11354},
  url={https://api.semanticscholar.org/CorpusID:237605338}
}

@InProceedings{khoa2023convolutional,
author="Doan, Nguyen Anh Khoa
and Racca, Alberto
and Magri, Luca",
title="Convolutional Autoencoder for the Spatiotemporal Latent Representation of Turbulence",
booktitle="Computational Science -- ICCS 2023",
year="2023",
publisher="Springer Nature Switzerland",
pages="328--335",
isbn="978-3-031-36027-5"
}

@article{solera2024beta,
  title={$\beta$-variational autoencoders and transformers for reduced-order modelling of fluid flows},
  author={Solera-Rico, Alberto and Sanmiguel Vila, Carlos and G{\'o}mez-L{\'o}pez, Miguel and Wang, Yuning and Almashjary, Abdulrahman and Dawson, Scott TM and Vinuesa, Ricardo},
  journal={Nature Communications},
  volume={15},
  number={1},
  pages={1361},
  year={2024},
  publisher={Nature Publishing Group UK London}
}

@article{schmid2010dynamic,
  title={Dynamic mode decomposition of numerical and experimental data},
  author={Schmid, Peter J},
  journal={Journal of Fluid Mechanics},
  volume={656},
  pages={5--28},
  year={2010},
  publisher={Cambridge University Press}
}

@article{abdi2010principal,
  title={Principal component analysis},
  author={Abdi, Herv{\'e} and Williams, Lynne J},
  journal={Wiley Interdisciplinary Reviews: Computational Statistics},
  volume={2},
  number={4},
  pages={433--459},
  year={2010},
  publisher={Wiley Online Library}
}

@incollection{lumley1967structure,
  title={The structure of inhomogeneous turbulent flows},
  author={Lumley, John L.},
  booktitle={Atmospheric Turbulence and Radio Wave Propagation},
  editor={Yaglom, A. M. and Tatarski, V. I.},
  pages={166--177},
  year={1967},
  publisher={Nauka, Moscow}
}

@article{bolla2021block,
  title={Block circulant matrices and the spectra of multivariate stationary sequences},
  author={Bolla, Marianna and Szabados, Tam{\'a}s and Baranyi, M{\'a}t{\'e} and Abdelkhalek, Fatma},
  journal={Special Matrices},
  volume={9},
  number={1},
  pages={36--51},
  year={2021},
  publisher={De Gruyter},
  doi={10.1515/spma-2020-0121}
}

@article{eyink2024onsager,
  title={Onsager's ‘ideal turbulence’theory},
  author={Eyink, Gregory},
  journal={Journal of Fluid Mechanics},
  volume={988},
  pages={P1},
  year={2024},
  publisher={Cambridge University Press}
}

@article{FukamiTaira2023,
  author = {Kai Fukami and Kunihiko Taira},
  title = {Grasping Extreme Aerodynamics on a Low-Dimensional Manifold},
  journal = {Nature Communications},
  volume = {14},
  number = {1},
  pages = {1--11},
  year = {2023},
}

@article{fukami2024single,
  title={Single-snapshot machine learning for super-resolution of turbulence},
  author={Fukami, Kai and Taira, Kunihiko},
  journal={Journal of Fluid Mechanics},
  volume={1001},
  pages={A32},
  year={2024},
  publisher={Cambridge University Press}
}

@article{Johnson2020,
  author = {Perry Johnson},
  title = {Energy Transfer from Large to Small Scales in Turbulence by Multiscale Nonlinear Strain and Vorticity Interactions},
  journal = {Physical Review Letters},
  volume = {124},
  number = {10},
  pages = {104501},
  year = {2020},
}

@article{Thomas2018,
  author = {Nathaniel Thomas and Tess Smidt and Steven Kearnes and Lusann Yang and Li Li and Kai Kohlhoff and Patrick Riley},
  title = {Tensor Field Networks: Rotation- and Translation-Equivariant Neural Networks for 3D Point Clouds},
  journal = {arXiv preprint arXiv:1802.08219},
  year = {2018},
}

@inproceedings{worrall2017harmonic,
  title={Harmonic networks: Deep translation and rotation equivariance},
  author={Worrall, Daniel E and Garbin, Stephan J and Turmukhambetov, Daniyar and Brostow, Gabriel J},
  booktitle={Proceedings of the IEEE conference on computer vision and pattern recognition},
  pages={5028--5037},
  year={2017}
}

@article{Azencot2020,
  author = {Omri Azencot and Nathan Erichson and Michael Mahoney},
  title = {Forecasting Sequential Data Using Consistent Koopman Autoencoders},
  journal = {arXiv preprint arXiv:2006.02548},
  year = {2020},
}

@inproceedings{cohen2016group,
  title={Group equivariant convolutional networks},
  author={Cohen, Taco and Welling, Max},
  booktitle={International conference on machine learning},
  pages={2990--2999},
  year={2016},
  organization={PMLR}
}

@article{KingmaWelling2014,
  author = {Diederik P. Kingma and Max Welling},
  title = {Auto-Encoding Variational Bayes},
  journal = {arXiv preprint arXiv:1312.6114},
  year = {2014}}

@inproceedings{helwig2023group,
author = {Helwig, Jacob and Zhang, Xuan and Fu, Cong and Kurtin, Jerry and Wojtowytsch, Stephan and Ji, Shuiwang},
title = {Group equivariant fourier neural operators for partial differential equations},
year = {2023},
publisher = {JMLR.org},
booktitle = {Proceedings of the 40th International Conference on Machine Learning},
articleno = {525},
numpages = {24},
series = {ICML'23}
}

@article{jimenez2018coherent,
  title={Coherent structures in wall-bounded turbulence},
  author={Jim{\'e}nez, Javier},
  journal={Journal of Fluid Mechanics},
  volume={842},
  pages={P1},
  year={2018},
  publisher={Cambridge University Press}
}

@article{Li2024,
  author = {Zongyi Li and Dezhi Zhang Huang and Burigede Liu and Anima Anandkumar},
  title = {Fourier Neural Operator with Learned Deformations for PDEs on General Geometries},
  journal = {arXiv preprint arXiv:2207.05209},
  year = {2024}}

@article{li2023fourier,
  title={Fourier neural operator with learned deformations for pdes on general geometries},
  author={Li, Zongyi and Huang, Daniel Zhengyu and Liu, Burigede and Anandkumar, Anima},
  journal={Journal of Machine Learning Research},
  volume={24},
  number={388},
  pages={1--26},
  year={2023}
}

@inproceedings{wang2020towards,
  title={Towards physics-informed deep learning for turbulent flow prediction},
  author={Wang, Rui and Kashinath, Karthik and Mustafa, Mustafa and Albert, Adrian and Yu, Rose},
  booktitle={Proceedings of the 26th ACM SIGKDD international conference on knowledge discovery \& data mining},
  pages={1457--1466},
  year={2020}
}

@inproceedings{DinhKB14,
  author       = {Laurent Dinh and
                  David Krueger and
                  Yoshua Bengio},
  title        = {{NICE:} Non-linear Independent Components Estimation},
  booktitle    = {3rd International Conference on Learning Representations, {ICLR} 2015,
                  San Diego, CA, USA, May 7-9, 2015, Workshop Track Proceedings},
  year         = {2015},
  url          = {http://arxiv.org/abs/1410.8516},
}

@article{li2020fourier,
  title={Fourier neural operator for parametric partial differential equations},
  author={Li, Zongyi and Kovachki, Nikola and Azizzadenesheli, Kamyar and Liu, Burigede and Bhattacharya, Kaushik and Stuart, Andrew and Anandkumar, Anima},
  journal={arXiv preprint arXiv:2010.08895},
  year={2020}
}

@inproceedings{DinhSB17,
  author       = {Laurent Dinh and
                  Jascha Sohl{-}Dickstein and
                  Samy Bengio},
  title        = {Density estimation using Real {NVP}},
  booktitle    = {5th International Conference on Learning Representations, {ICLR} 2017,
                  Toulon, France, April 24-26, 2017, Conference Track Proceedings},
  publisher    = {OpenReview.net},
  year         = {2017},
}

@article{fuchs2020se,
  title={Se (3)-transformers: 3d roto-translation equivariant attention networks},
  author={Fuchs, Fabian and Worrall, Daniel and Fischer, Volker and Welling, Max},
  journal={Advances in neural information processing systems},
  volume={33},
  pages={1970--1981},
  year={2020}
}

@article{kingma2018glow,
  title={Glow: Generative flow with invertible 1x1 convolutions},
  author={Kingma, Durk P and Dhariwal, Prafulla},
  journal={Advances in neural information processing systems},
  volume={31},
  year={2018}
}

@inproceedings{behrmann2019invertible,
  title={Invertible residual networks},
  author={Behrmann, Jens and Grathwohl, Will and Chen, Ricky TQ and Duvenaud, David and Jacobsen, J{\"o}rn-Henrik},
  booktitle={International conference on machine learning},
  pages={573--582},
  year={2019},
  organization={PMLR}
}

@article{toth2019hamiltonian,
  title={Hamiltonian generative networks},
  author={Toth, Peter and Rezende, Danilo Jimenez and Jaegle, Andrew and Racani{\`e}re, S{\'e}bastien and Botev, Aleksandar and Higgins, Irina},
  journal={arXiv preprint arXiv:1909.13789},
  year={2019}
}
\end{document}